\documentclass{article}

\usepackage[preprint]{neurips_2025}

\usepackage{natbib}

\usepackage[utf8]{inputenc} % allow utf-8 input
\usepackage[T1]{fontenc}    % use 8-bit T1 fonts
\usepackage{hyperref}       % hyperlinks
\usepackage{url}            % simple URL typesetting
\usepackage{booktabs}       % professional-quality tables
\usepackage{amsfonts}       % blackboard math symbols
\usepackage{nicefrac}       % compact symbols for 1/2, etc.
\usepackage{microtype}      % microtypography
\usepackage{xcolor}         % colors

\usepackage{multirow}
\usepackage{multicol}
\usepackage{graphicx}
\usepackage{tabularx}
\usepackage{colortbl}
\usepackage{amsmath} 
\usepackage{wrapfig}

\definecolor{lightblue}{HTML}{E0ECFF}
\definecolor{lightred}{HTML}{FFD4DA}
\definecolor{platinum}{HTML}{E5E4E2}

\definecolor{ourgreen}{HTML}{F0FFF2}
\definecolor{ourred}{HTML}{FFF5F5}

\definecolor{promptgrey}{HTML}{F2F3F4}

\definecolor{mygreen}{RGB}{75, 225, 75}
\hypersetup{
    colorlinks=true,       % Enable colored links
    citecolor=mygreen,         % Color of citation links (e.g., \cite{})
}

\title{TokBench: Evaluating Your Visual Tokenizer before Visual Generation}

\vspace{-5pt}
\author{
  \vspace{-25pt}\\
  \textbf{Junfeng Wu\thanks{Equal contribution.} ,\quad Dongliang Luo$^{*}$,\quad Weizhi Zhao,\quad Zhihao Xie,} \\  \textbf{Yuanhao Wang, \quad  Junyi Li, \quad Xudong Xie,  \quad Yuliang Liu,\quad Xiang Bai
  }\vspace{3pt} \\
  Huazhong University of Science and Technology  \\
  \texttt{\small wjf5203@gmail.com, ldl@hust.edu.cn, zhaoweizhi@hust.edu.cn, } \\ 
  \texttt{\small zhxie17@hust.edu.cn, yhwang7@hust.edu.cn, ljy1308598378@gmail.com,  }  \\
  \texttt{\small xdxie@hust.edu.cn, ylliu@hust.edu.cn, xbai@hust.edu.cn}\vspace{8pt}  \\
  \vspace{-7pt} \\
}

\begin{document}

\maketitle
\begin{figure}[ht]
    \vspace{-.4in}
    \centering
    \raisebox{-0.3em}{\includegraphics[height=1.5em]{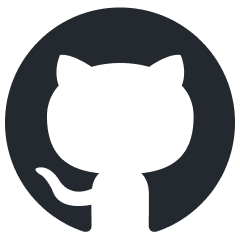}} \hspace{-.1em} HomePage: \hspace{.05in} \url{https://wjf5203.github.io/TokBench}\\
    \raisebox{-0.6em}{\includegraphics[height=2em]{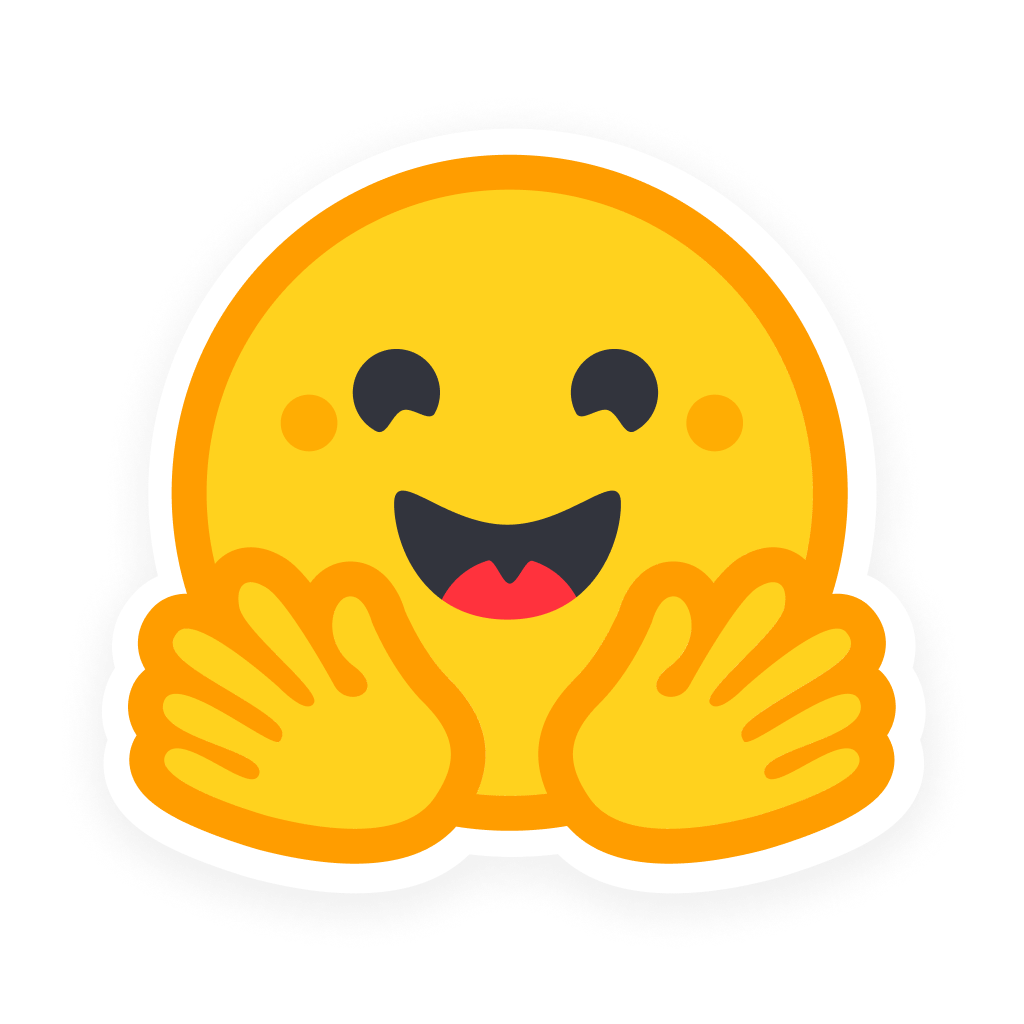}} \hspace{-.1em} Dataset: \hspace{.05in} \url{https://huggingface.co/datasets/Junfeng5/TokBench}
\end{figure}

\begin{abstract}

In this work, we reveal the limitations of visual tokenizers and VAEs in preserving fine-grained features, and propose a benchmark to evaluate reconstruction performance for two challenging visual contents: text and face. Visual tokenizers and VAEs have significantly advanced visual generation and multimodal modeling by providing more efficient compressed or quantized image representations. However, while helping production models reduce computational burdens, the information loss from image compression fundamentally limits the upper bound of visual generation quality. To evaluate this upper bound, we focus on assessing reconstructed text and facial features since they typically: 1) exist at smaller scales, 2) contain dense and rich textures, 3) are prone to collapse, and 4) are highly sensitive to human vision. We first collect and curate a diverse set of clear text and face images from existing datasets. Unlike approaches using VLM models, we employ established OCR and face recognition models for evaluation, ensuring accuracy while maintaining an exceptionally lightweight assessment process \textbf{requiring just 2GB memory and 4 minutes to complete.} Using our benchmark, we analyze text and face reconstruction quality across various scales for different image tokenizers and VAEs. Our results show modern visual tokenizers still struggle to preserve fine-grained features, especially at smaller scales. We further extend this evaluation framework to video, conducting comprehensive analysis of video tokenizers. Additionally, we demonstrate that traditional metrics fail to accurately reflect reconstruction performance for faces and text, while our proposed metrics serve as an effective complement.

\end{abstract}

\section{Introduction}

\begin{figure}
  \centering
    \includegraphics[width=0.98 \linewidth]{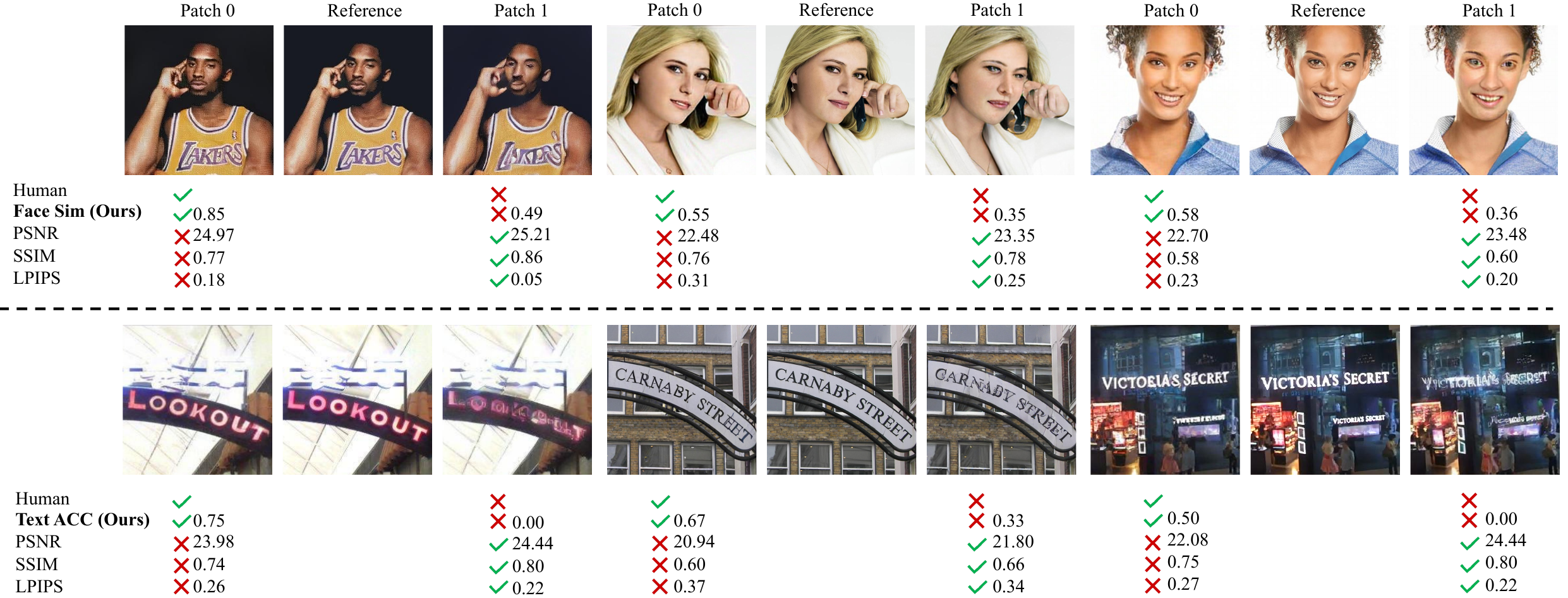}
  \vspace{-2mm}
  \caption{\textbf{Comparison of Different Metrics with Human Judgments.} In each case, previous metrics (PSNR, SSIM, LPIPS) demonstrate discrepancies with human assessments, whereas our proposed face similarity and text accuracy effectively reflect the reconstruction quality. The reference image represents the original, while Patch 0 and Patch 1 show reconstruction results from different visual tokenizers. The same regions are cropped from the complete images for visualization.}
  \label{fig:metric_compare}
  \vspace{-4mm}
\end{figure}

In recent years, we have witnessed rapid advancements in visual generation and its tremendous application potential. Diffusion models~\cite{ldm,sdxl,sd3,DiT,flux2024} have elevated the quality of visual generation to amazing levels while enabling versatile conditional control. Meanwhile, autoregressive approaches~\cite{DALLE,llamagen,team2024chameleon,wu2024liquid} have gradually demonstrated comparable performance and the potential for seamless integration with large language models (LLMs), offering a unified framework for multimodal generation.

Early diffusion models~\cite{ho2020denoising,song2021denoising} operated directly in pixel space, but their high computational cost motivated subsequent works~\cite{ldm, sdxl, DiT} to shift the diffusion process into the latent space of pretrained variational autoencoders (VAEs)~\cite{VAE, ldm}. This approach achieves a near-optimal trade-off between computational efficiency and detail preservation.
In contrast to diffusion-based methods, which decompose image generation into iterative denoising steps, autoregressive models~\cite{esser2021taming, DALLE} generate visual content sequentially while achieving comparable or even superior~\cite{llamagen,var} visual quality. Their inherent compatibility with LLMs further positions them as promising candidates for unified multimodal generation frameworks~\cite{liu2024world, team2024chameleon, wu2024liquid}.
For autoregressive visual generation, VQVAE~\cite{van2017neural} first introduced discrete latent representations of images, modeling their distribution autoregressively. VQGAN~\cite{esser2021taming} significantly improved reconstruction quality, enabling efficient high-resolution image synthesis via transformers or LLMs. 
Both image generation approaches have been successfully extended to the video generation domain~\cite{hong2023cogvideo,yang2024cogvideox,kong2024hunyuanvideo,ma2025step}.
However, encoding images or videos into latent space typically incurs information loss, particularly due to vector quantization (VQ) from continuous features to discrete tokens. This loss fundamentally constrains the upper bound of generation fidelity.

% The field continues to rely primarily on conventional metrics such as PSNR, SSIM, and FID, which, while well-established, exhibit known limitations in capturing perceptual fidelity and semantic consistency - particularly for modern high-resolution, multimodal generation scenarios. This evaluation gap becomes increasingly consequential as tokenizer designs grow more sophisticated, underscoring the need for novel assessment frameworks that can better align with human perceptual judgments and downstream task performance.

There have been several classical methods for evaluating the quality of reconstructed images. Traditional pixel-level metrics, such as PSNR, measure pixel-wise intensity differences, emphasizing global fidelity but disregarding perceptual relevance. SSIM~\cite{SSIM} and FSIM~\cite{FSIM} further incorporate luminance, contrast, structural, and edge-texture information, but they are more sensitive to noise. These pixel-level metrics typically focus on only few aspects of image quality and fail to measure similarity in a way that aligns with human judgment. To address these limitations, feature-based metrics like FID~\cite{FID}, IS~\cite{InceptionScore}, and LPIPS~\cite{LPIPS} have emerged to assess semantic and distributional consistency of reconstructed images using features from pretrained networks. While these feature-based metrics better approximate human perception compared to pixel-level ones, their reliance on pretrained models makes evaluation unreliable when reconstructed images deviate from the pretraining distribution, as illustrated in Fig~\ref{fig:metric_compare}.

Since human judgments of similarity depend on high-order, context-dependent image structures that may not conform to feature distance metrics, we naturally consider certain high-dimensional image features - particularly faces and texts - are more reliant on human assessment than generic natural image characteristics. Compared to other visual contents, the detection and evaluation of faces and text have been extensively studied, resulting in mature toolchains~\cite{doctr2021,insightface}. Moreover, unlike subtle pixel-level variations, \textbf{text readability} and \textbf{identity preservation} are far more perceptually critical to human observers. Pixel-level metrics fail to penalize semantically critical errors (e.g., misaligned strokes in text), while feature-based metrics lack the granularity to assess domain-specific attributes (e.g., facial symmetry or character recognition accuracy). This gap highlights the need for a tailored benchmark that integrates task-aware evaluation to complement existing metrics.

To address this gap, we propose Visual Tokenizer Benchmark (TokBench).
Specifically, we curated 12,398 images and 403 video clips (51,590 frames) rich in faces and text from publicly available datasets, encompassing both natural scenes and document contexts, with balanced scale distributions for both facial and text content.
To assess text reconstruction quality, we employ an OCR model to determine whether the reconstructed text remains accurately recognizable, subsequently computing the T-ACC (Text Recognition Accuracy) and T-NED (Text Normalized Edit Distance) metrics. For facial content, we leverage a face recognition model to extract facial features and compute the F-Sim (Facial Similarity) metric, quantifying identity preservation. 
For reconstructed videos, we perform a frame-by-frame evaluation and report the average results.
These metrics offer intuitive quantification of a visual tokenizer’s ability to retain the most visually challenging content types—areas where current evaluation methods frequently underperform. Leveraging this benchmark, we conducted a comprehensive evaluation of existing visual tokenizers and VAEs, demonstrating that the proposed metrics serve as a meaningful complement to conventional reconstruction quality standards.

In summary, the main contributions of this paper can be categorized into the following points:
\begin{itemize}
    \item {We reveal that conventional metrics exhibit inconsistencies with human evaluation when assessing the reconstruction quality of human-sensitive content like text and face.}
    \item {We propose TokBench, comprising a diverse image dataset rich in faces and text, along with a lightweight evaluation pipeline, \textbf{requiring only 2GB VRAM within 4 minutes}.}
    \item {We conduct comprehensive evaluations of existing image tokenizers and VAEs on face and text reconstruction, and further extend this assessment to video tokenizers to explore the upper bounds of visual generation models.}

\end{itemize}

% In contrast to conventional reconstruction evaluation methods, we propose a more direct assessment of two high-frequency yet perceptually critical visual elements: text and facial features. The fundamental challenge mirrors the adage that "one cannot accurately depict the Mona Lisa without having seen her" — if an image tokenizer fails to preserve these sensitive details during compression, the lost information may introduce noise during generative model training, ultimately limiting both learning efficiency and performance ceilings. To address this gap, we introduce three novel evaluation metrics (T-ACC, T-NED, and F-Sim) designed to complement existing reconstruction assessments. These metrics provide intuitive quantification of a tokenizer's capability to preserve the most challenging visual content types, where current methods often fall short.
% Experiments xxxxx
\section{Related Work}
\subsection{Visual Tokenizers and VAEs}
\paragraph{Image} Since Latent Diffusion Models~\cite{ldm} achieved promising results by learning visual generation in VAE's latent space, the study of continuous or discrete visual latent spaces has played a critical role in visual generation, with increasing exploration focused on tokenizer design. The conventional VAE~\cite{dai2019diagnosing, VAE} demonstrated both theoretical and empirical evidence for the advantages of learning a data representation encoded to images with a learned generator. \cite{van2017neural} introduced the Vector Quantised Variational Autoencoder (VQVAE), which learns discrete representations of images and models their distribution autoregressively. VQGAN~\cite{esser2021taming} further enhances the visual reconstruction capability of VQVAE by incorporating GAN loss and demonstrates the potential of autoregressive models in generating high-resolution images. Visual AutoRegressive modeling (VAR)~\cite{var} redefined autoregressive learning on images as a coarse-to-fine next-scale prediction.
% showing superior generalization and scaling capabilities compared to diffusion transformers while requiring fewer steps.
UniTok~\cite{ma2025unitok} explores the introduction of semantic informations training for discrete visual tokens, enriching semantic information to further improve the understanding and generation capabilities of unified models~\cite{team2024chameleon, wu2024liquid}. Meanwhile, VAVAE~\cite{yao2025vavae} and REPA~\cite{yu2025repa} address the high-dimensional challenges of continuous VAE spaces by leveraging semantic space supervision, while TokenBridge~\cite{wang2025bridging} and Layton~\cite{xie2025layton} explore the communication and fusion between continuous and discrete tokens.
In a different vein, MAGVIT-v2~\cite{magvit2}, FSQ~\cite{fsq}, BSQViT~\cite{bsq} propose lookup-free quantization, presenting an alternative approach that bypasses traditional lookup mechanisms. 
TiTok~\cite{titok} performs 2D-to-1D distillation, compressing the number of tokens used to represent the same image.
% Despite the flourishing innovation in visual tokenizers, the evaluation methods for visual reconstruction quality remain unchanged, relying mainly on rFID or LPIPS fails to fully reflect human visual assessment standards.

\vspace{-2mm}
\paragraph{Video}
Videos contain both spatial and temporal information, making their data volume substantially larger than images. Early video models typically employed image VAEs or VQVAEs~\cite{hong2023cogvideo} directly for generation, but spatial-only modeling often produces jittery outputs. Some approaches~\cite{lin2024open,zheng2024opensora} attempted 3D VAEs for temporal compression, yet limited latent channels still yielded blurry and unstable results. Recent methods~\cite{ma2025step,kong2024hunyuanvideo,yang2024cogvideox} utilizing 3D Causal VAEs have demonstrated superior video encoding performance.

\subsection{Evaluation of Image Reconstruction}

\paragraph{Pixel-level Evaluation}
Traditional low-level metrics assess reconstruction quality through pixel-wise comparisons. Mean Squared Error (MSE) quantifies average squared intensity differences, while Peak Signal-to-Noise Ratio (PSNR) extends this concept logarithmically using the ratio between the maximum possible power of a signal and the power of corrupting noise that affects the fidelity of its representation. 
% The PSNR is frequently used to evaluate the performance of several image processing methods, including compressors, filters, and related apparatus. A higher PSNR value in this case denotes a more effective compression or filtering method for maintaining image quality. 
The structural similarity index measure (SSIM)~\cite{SSIM} models human perception through luminance, contrast, and structural comparison, carrying important information about the structure of the objects in the visual scene. Feature Similarity Index (FSIM)~\cite{FSIM} measures the similarity between two images based on their low-level features. HDR-VDP~\cite{HDRVDP} specializes in varying luminance conditions, predicting both quality degradation and change visibility.
\vspace{-2mm}
\paragraph{Feature-level Evaluation}
Previous pixel-level metrics are simple, shallow functions, and fail to account for many nuances of human perception. Advanced feature-level metrics leverage deep learning for semantic evaluation. Learned Perceptual Image Patch Similarity (LPIPS)~\cite{LPIPS} compares deep features from pretrained networks to better align with human judgment. Fréchet Inception Distance (FID)~\cite{FID} measures distributional similarity between generated and real images using Inception-v3 features, while Inception Score (IS)~\cite{InceptionScore} evaluates both diversity and recognizability through classifier predictions. These high-level metrics address limitations of pixel-based methods but require careful interpretation when evaluating out-of-distribution samples. 
% Furthermore, although feature-level evaluations can approximate human judgment of overall image quality, 
Furthermore, these features typically represent high-dimensional global characteristics, small-scale objects such as text and faces have a relatively minor influence on these features. As illustrated in Figure~\ref{fig:metric_compare}, previous metrics fail to reflect the reconstruction quality of small-scale objects, which is a critical aspect that modern high-quality visual generation models particularly focus on.

\subsection{Text and Face Datasets}

\paragraph{Text Data}
Texts are representative texture elements in images and unsatisfactory generation quality would seriously affect their readability. 
% To evaluate this, we naturally convert the assessment of reconstructed appearance into a text recognition problem. 
Previous datasets for text recognition are focused on cropped text regions, restricting the diversity of text scales and image scenarios. Therefore, we consider collecting data from text spotting datasets~\cite{ic13,ic15,totaltext,textocr,ctw1500}, which are annotated with the locations and transcriptions of texts. Additionally, some datasets for key information extraction~\cite{sroie,cord} and document-oriented VQA~\cite{docvqa,infographicvqa} also provide the above annotations. In this work, we collect text data from 8 different text image datasets that vary in fonts, styles, scales and backgrounds, enriching the comprehensiveness of our benchmark. In addition, text spotting in videos has been receiving growing attention recently, and the related datasets~\cite{ic15,wu2024dstext} are released. They support us to further extend our assessment to video tokenizers. 
We unify the text representations for consistent evaluation.
% Therefore, we integrated and processed text datasets from multiple sources, covering scenarios including scene text, document and graphic. 
\vspace{-2mm}
\paragraph{Face Data}
% zwz
For evaluating face generation quality, we considered datasets originally curated for two primary face-related tasks: facial landmark detection and face recognition. Key datasets for facial landmark detection include WFLW~\cite{WFLW}, 300W~\cite{300W}, and AFLW~\cite{AFLW}. For face recognition, frequently utilized datasets include LFW~\cite{LFW}, CALFW~\cite{CALFW}, and CFPW~\cite{CFPW}, among others. However, most of these datasets were deemed unsuitable for our benchmark since they consist predominantly of single-face portrait images, which do not accurately represent the distribution of faces in “in-the-wild” scenarios.
Consequently, we selected the WFLW dataset, which composed of images captured in naturalistic, unconstrained environments, which often contain multiple faces. 
For video data, we observe that many video understanding datasets contain abundant scenes and faces. For instance, VideoMME~\cite{videomme}, MVBench~\cite{li2024mvbench}, and MMBench-Video~\cite{fang2024mmbenchvideo} are popular benchmarks for evaluating multimodal video understanding in VLLMs, which include numerous facial segments that can serve as our data pool.

% It comprises 6,551 images, a quantity we considered appropriate for our evaluation purposes. Furthermore, its ground truth, when augmented by detections from a face detector, provides a total of 17,700 face annotations. This scale is comparable to that of our OCR datasets.

\section{TokBench}

Our goal is to provide a novel benchmark specifically designed to evaluate the reconstruction quality of two critical visual elements: texts and human faces in images. To establish this benchmark, we first curate a diverse collection of images rich in textual and facial content, systematically categorized by their spatial scales within the images. 
% We conducted comprehensive analyses of the distribution of these reconstruction targets across different scales and their corresponding recognition difficulties. 
Then we incorporate specialized evaluation metrics that assess: (1) the legibility of reconstructed text and (2) identity preservation in reconstructed faces. 
As a result, TokBench provides targeted evaluation of discrete or continuous tokenizers' capability in reconstructing faces and text, thereby ensuring the upper bound of high-quality visual generation.
Furthermore, we curate videos containing rich texts and faces to extend TokBench to assess video tokenizers and VAEs.

\begin{figure}
  \centering
  \includegraphics[width=0.9 \linewidth]{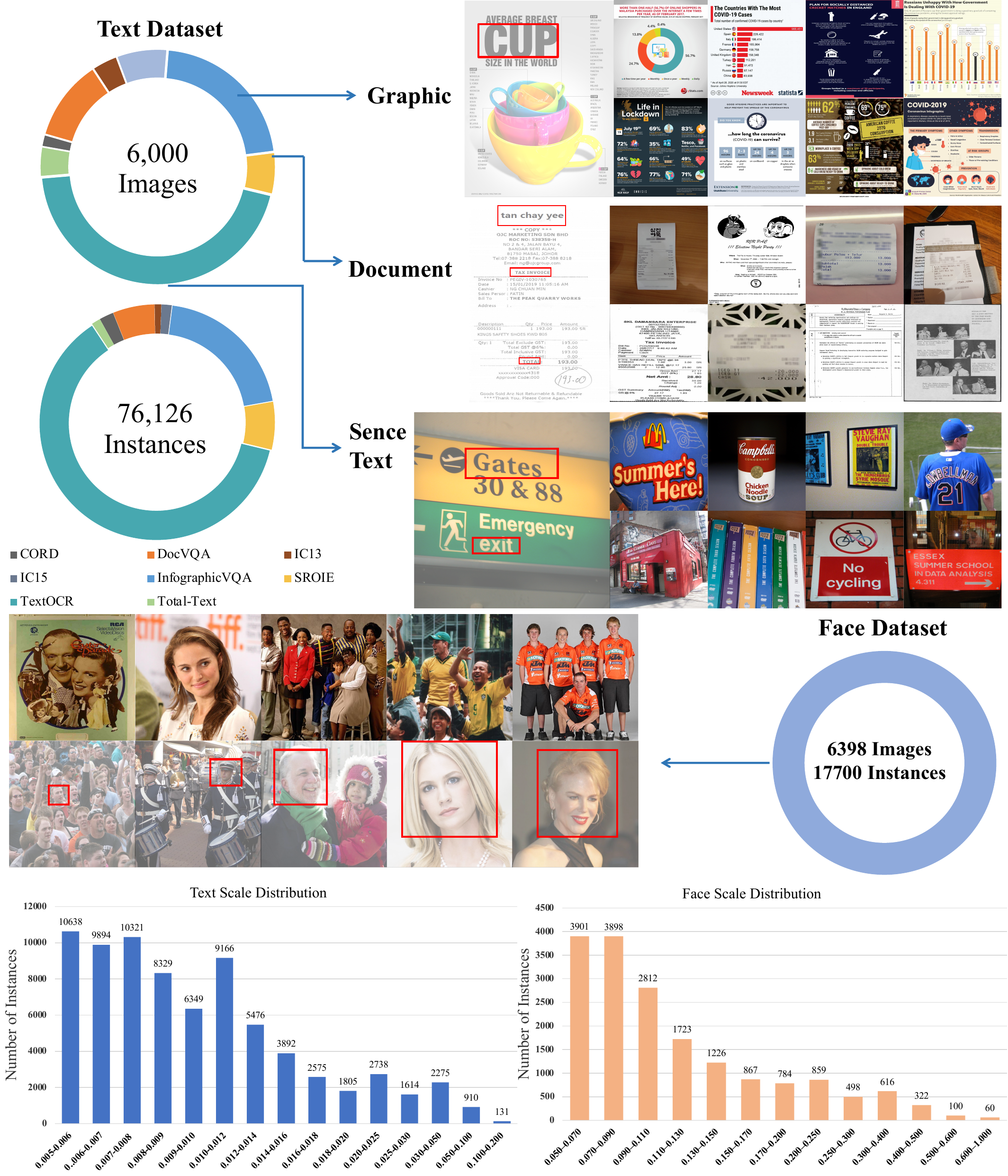}
  \vspace{-2mm}
  \caption{\textbf{Statistics and Sample Diversity of TokBench-Image.} TokBench features a balanced instance-scale distribution with particular emphasis on small-scale face and text instances, presenting significant challenges for existing visual reconstruction approaches.}
  \label{fig:dataset}
  \vspace{-4mm}
\end{figure}

\subsection{Image Data Curation}

\subsubsection{Text Data Curation}
\label{sec:text}

\paragraph{Data Collection}
We first collect text images from eight existing open-source datasets for diversity. Specifically, they include scene text datasets, \textit{i.e.}, ICDAR 2013~\cite{ic13}, IC15~\cite{ic15}, Total-Text~\cite{totaltext} and TextOCR~\cite{textocr}, and document datasets, \textit{i.e.}, CORD~\cite{cord}, SROIE~\cite{sroie}, InfographicVQA~\cite{infographicvqa} and DocVQA~\cite{docvqa}. We use their validation or accessible test set to build our benchmark. For datasets that are not divided into training and test sets, we sample from them. These datasets provide word-level annotations that contain both the position and transcription for each text instance, allowing us to perform consistent evaluations. Next, we uniformly use the horizontal bounding box $\{x_i^t, y_i^t, w_i^t, h_i^t\}$ to represent the the $i$-th text regions.

\vspace{-2mm}
\paragraph{Difficulty Rating}
We consider the relative scale of texts as the major factor distinguishing the reconstruction difficulty of the evaluated data.
Due to the large variation of scales and character lengths of texts, we focus on the character-level text scale for measurement, which can be approximately derived from annotations.
Given a text image $I^t \in \mathbb{R}^{H \times W \times 3}$. We assume that characters are uniformly distributed in the bounding box for most texts. Thus, we approximate the relative scale of the $i$-th text by normalizing the scale of one character by the maximum length of the image:

\begin{equation}
    r_i^t = \frac{max(h_i^t, w_i^t)}{max(H_i, W_i) \times N_i^c},
\end{equation}

where $N^i_c$ is the number of characters of the $i$-th text instance.

\begin{figure}[tb]
  \centering
  \includegraphics[width=0.9 \linewidth]{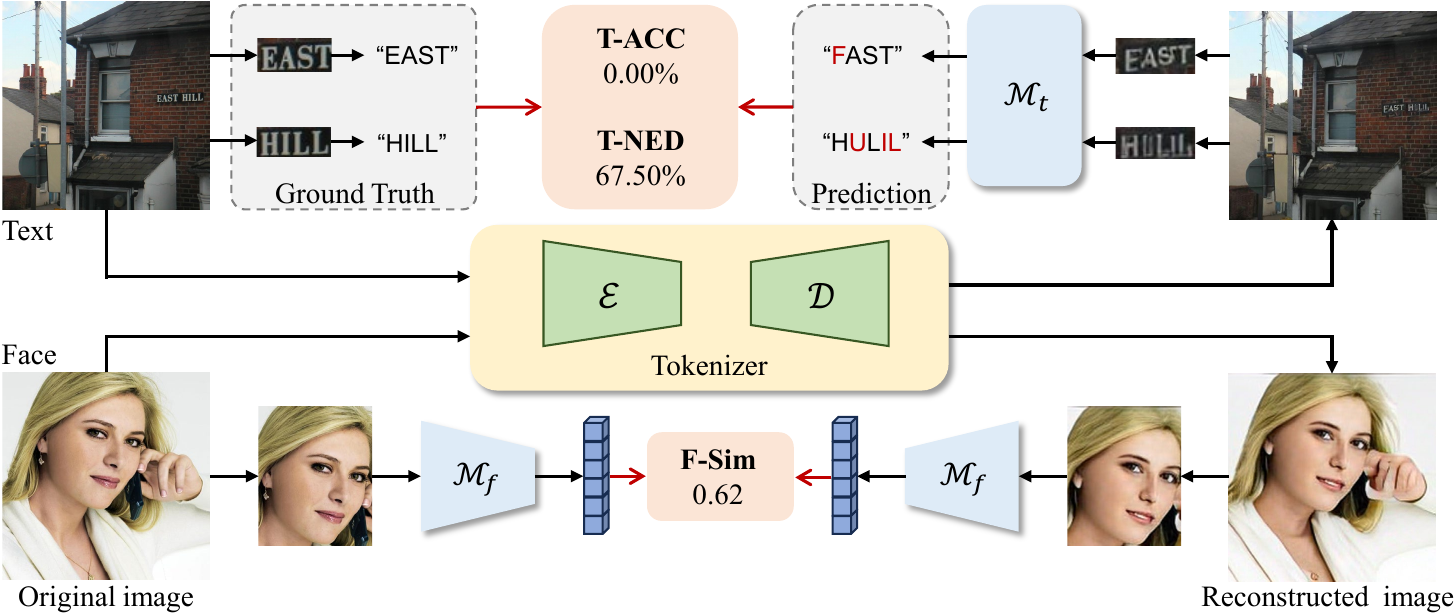}
  \vspace{-2mm}
  \caption{Overview of the evaluation process of TokBench.}
  \label{fig:pipeline}
  \vspace{-2mm}
\end{figure}

\begin{figure}[tb]
  \centering
  \includegraphics[width=1.0 \linewidth]{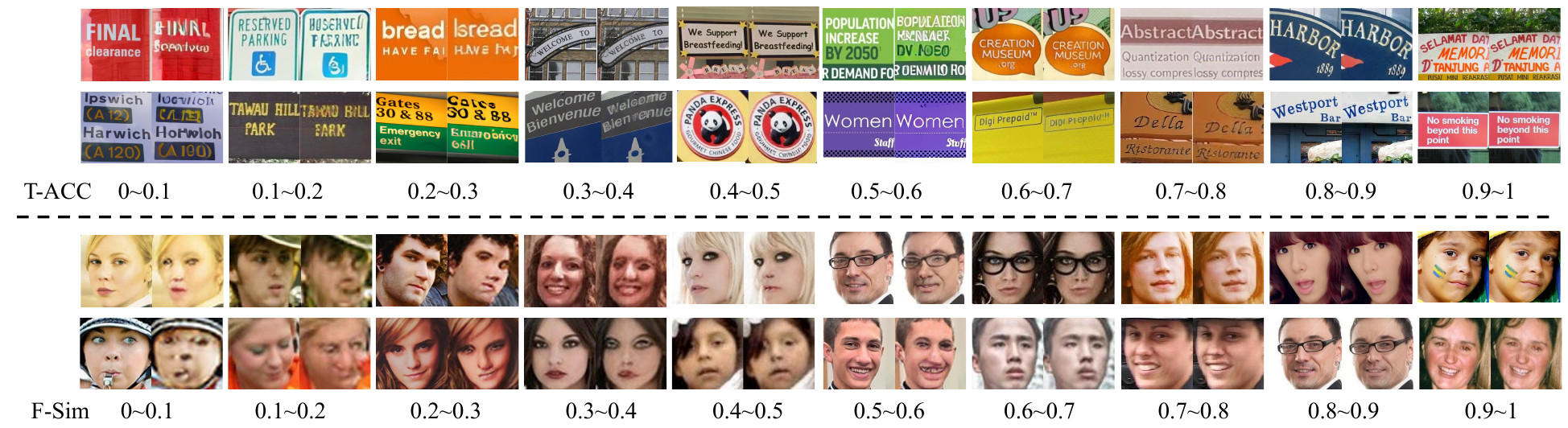}
  \vspace{-4mm}
  \caption{Comparison between reconstructed images (right) and original images (left) under different T-ACC and F-Sim metrics. Higher metric values indicate reconstructed images that more closely resemble the original. (Zoom in for better comparison.)}
  \label{fig:metric_visualize}
    \vspace{-2mm}
\end{figure}

\paragraph{Data Cleaning}
The feasibility of reconstructing tiny regions should be considered. Meanwhile, the assessment of the reconstruction quality of text images is based on a pretrained text recognition model $\mathcal{M}_t$, requiring the predictions of $\mathcal{M}_t$ completely accurate on the original images. To ensure the validity of the evaluation, we remove extremely tiny cases and unrecognized instances that would cause ambiguity with the following steps:
1) We assume the minimum pixels to clearly represent a character is $5\times5$. Hence, we remove instances with $min(h^t, w^t) < 5$ or $r^t < 0.005$.
2) We filter out the instances containing characters out of the vocabulary of the recognizer and regions that contain only one special symbol, avoiding ambiguous and invalid recognition results.
3) We only keep text instances that can be correctly recognized by $\mathcal{M}_t$ from the remaining, guaranteeing the performance degradation in the benchmark is mainly caused by poor reconstructions.
Afterward, we keep the images that contain at least one valid text instance.

As a result, the text set in TokBench consists of 6,000 images and 76,126 valid text instances as shown in Fig.~\ref{fig:dataset}. Multiple sources enrich the diversity of text fonts, styles, scales and backgrounds. Each instances is annotated using $\{x_i^t, y_i^t, w_i^t, h_i^t, r_i^t, \hat{s}_i\}$, where $\hat{s}_i$ is the ground truth transcription. Using $r_i^t$, we empirically set 3 different difficulty levels (Small, Medium, and Large). The lowest limit scale in evaluation for the resolution $L$ during reconstruction is no less than $5/L$, so that the text regions are valid as illustrated in data cleaning. The scale range for each level is in the Appendix.

\subsubsection{Face Data Curation}
\vspace{-2mm}
For our facial data source, we select WFLW~\cite{WFLW} due to its uniform distribution of face scales and diverse scenarios. From the original 6,551 images, we first filter out all images with aspect ratios exceeding 2, retaining 6,398 valid images containing 9,739 ground-truth (GT) annotated face instances. Since many images contained unannotated faces, we perform additional face detection using the antelopev2 model from insightface~\cite{insightface}, keeping only detections with confidence scores above 0.5. For the detected faces, we calculate each face's scale by dividing the longer side of the bounding box by the image, retaining only faces with scales greater than 0.05 as supplementary GT data. This process yields 17,700 valid target faces, on which we will evaluate the similarity between reconstructed faces and original facial features.

\subsection{Evaluation Protocols}
\vspace{-2mm}

The overall evaluation pipeline is illustrated in Fig.~\ref{fig:pipeline}. Text and face images are first reconstructed by the given visual tokenizer $\mathcal{T}$.
For the reconstructed text images, each valid text region is cropped according to the ground truth (GT). The cropped regions are fed into a pretrained text recognition model $\mathcal{M}_t$, obtaining the transcription predictions, which are further evaluated by the corresponding GT using T-ACC and T-NED metrics.
Similarly, for the face images, each face area is cropped by GT. The corresponding areas between the original image and the reconstructed image are encoded by a pretrained face recognition model $\mathcal{M}_f$. The encoded feature vectors are measured by F-Sim to evaluate the quality of the generated face.

\paragraph{Text}
We choose the recent PARSeq~\cite{parseq} as the pretrained recognizer for its good balance between accuracy and efficiency. We use the implementation by docTR~\footnote{\url{https://github.com/mindee/doctr}}~\cite{doctr2021}, an OCR toolbox which can be easily installed.
Following the metrics in text recognition tasks, the results are evaluated by the text recognition accuracy (T-ACC) and Normalized Edit Distance (T-NED)~\cite{zhang2019icdar} between the recognition result $s_i$ and the ground truth $\hat{s}_i$. Since our goal is to assess the reconstruction quality, we distinguish between uppercase and lowercase letters because their appearances are different, which should be maintained after a decent reconstruction. It is regarded as a true positive only when the predicted word is exactly the same as GT in our T-ACC metric. Secondly, T-NED gives a more fine-grained analysis considering the accuracy of characters, which is formulated as:
% Accuracy(ACC) and 1-NED
\begin{equation}
    \text{T-NED} = 1 - \sum_i^{N^t}\frac{D(s_i, \hat{s}_i)}{max(l_{i}, \hat{l}_{i})},
\end{equation}
where $l_i$ and $\hat{l}_i$ are the numbers of characters of the predicted text and the corresponding GT. $N^t$ is the number of text instances. $D$ indicates the Levenshtein distance.

\paragraph{Face}
Just as one cannot paint the Mona Lisa without having seen her, a visual tokenizer that fails to accurately reconstruct faces will prevent generative models trained on its latent space from correctly generating corresponding identities. In fact, distorted identities may even mislead the learning process of generative models. To evaluate the fidelity of face reconstruction, we employ the insightface~\cite{insightface} recognition model $\mathcal{M}_f$ to measure the similarity between reconstructed and original faces. Specifically, we input the same facial keypoints from annotations with both original and reconstructed images into the recognition model to extract corresponding facial features, then compute the cosine distance between these feature vectors as our face similarity metric (F-Sim). As shown in Figure~\ref{fig:metric_visualize}, higher similarity scores indicate better face reconstruction quality, with Table 1 in Supp. demonstrating that high-resolution resizing achieves the highest F-Sim of 1.
% In Table~\ref{fig:metric_visualize}, we visualize the reconstruction quality corresponding to different T-ACC and F-Sim scores.

\subsection{Video Data Curation}
\paragraph{Text}
We collect real-world videos from the ICDAR 2013–15 Text-in-Videos Challenge~\cite{ic15} and the test set of DSTextV2~\cite{wu2024dstext}. Word-level annotations for texts in each frame are given. Similar to the processing procedures illustrated in Sec.~\ref{sec:text}, we get rid of invalid text instances while preserving the original video clips. Since the resizing strategy for video tokenizers is based on the short side, we remove instances with $min(h^t, w^t) < 5$ or $r^t < \frac{5 \times min(H, W)}{480\times max(H, W)}$, where 480 is the upper bound of resized short side in our evaluation. Thus, we obtained 15,921 frames that contain 347,468 valid text instances. The evaluation is conducted per frame, whose pipeline and metrics are consistent with Fig.~\ref{fig:pipeline}. We only need to recognize text in the cropped regions while ignoring frames containing no valid text, improving the efficiency.

\paragraph{Face}
We first downloaded all videos from the VideoMME~\cite{videomme}, MVBench~\cite{li2024mvbench}, and MMBench-Video~\cite{fang2024mmbenchvideo} datasets. Each video was sampled at 1 FPS and processed using insightface~\cite{insightface} for face detection, retaining only videos containing faces with the longer edge exceeding 512 pixels. The retained videos then underwent frame-by-frame analysis to select clips meeting two criteria: continuous face presence for at least 3 seconds and detection of more than 3 faces. After filtering out videos where most frames contained only a single face, we manually curated the remaining clips based on video quality and content richness, resulting in 328 selected 3-second video segments (25,980 frames total). Within these frames, we performed additional insightface detection to identify faces with confidence scores above 0.5 and scale factors exceeding 0.03, yielding 81,556 valid target faces for frame-by-frame similarity evaluation between reconstructed and original faces.

\section{Experiments}

\begin{table*}[!t]

\centering

\small
\setlength{\tabcolsep}{4pt}
\resizebox{\textwidth}{!}{\begin{tabular}{l l  c c c c c c c  |c c c c }

\toprule

 \multirow{2}{*}{\textbf{Type}} & \multirow{2}{*}{\textbf{Method}} & \multirow{2}{*}{\textbf{Factor}} & \multicolumn{4}{c}{{\bf Text}(\%)} & \multicolumn{2}{c}{\bf Face}  & \multirow{2}{*}{\bf rFID{$\downarrow$}} & \multirow{2}{*}{\bf LPIPS{$\downarrow$}} & \multirow{2}{*}{\bf PSNR{$\uparrow$}} & \multirow{2}{*}{\bf SSIM{$\uparrow$}} \\
 
\cmidrule(lr){4-7} \cmidrule(lr){8-9}
& & &{T-ACC$_{s}\uparrow$} &{T-ACC$_{m}\uparrow$} &{T-NED$_{s}\uparrow$}  &{T-NED$_{m}\uparrow$}&{F-Sim$_{s}\uparrow$} &{F-Sim$_{m}\uparrow$}   &&&& \\

\midrule
\multicolumn{13}{c}{\textit{Resolution: 256 $\times$ 256}} \\
\midrule
\cellcolor{lightblue} &\cellcolor{lightblue}Resize &\cellcolor{lightblue}$1\times$ &\cellcolor{lightblue}86.05    &\cellcolor{lightblue}93.02  &\cellcolor{lightblue} 92.98     &\cellcolor{lightblue}96.53 &\cellcolor{lightblue} 0.85 &\cellcolor{lightblue}0.93 &\cellcolor{lightblue}5.39 &\cellcolor{lightblue}0.06 &\cellcolor{lightblue}27.71 &\cellcolor{lightblue}0.84\\

\multirow{14}{*}{Discrete} 
&TiTok~\cite{titok} &$1D$ &0.05 &0.09 &3.04 &4.23   &0.03 &0.04  &16.25 &0.52 &13.54 &0.47  \\
&FlexTok~\cite{flextok} &$1D$  &0.55   &6.95 &7.80  &21.09  &0.06  &0.15 & 8.87 & 0.35 &17.37 &0.57 \\
&VQGAN~\cite{esser2021taming} &$16\times$ &0.05&1.10&4.34&8.22&0.05 &0.10 &12.63 &0.36 &17.29 &0.55  \\ 
&Chameleon~\cite{team2024chameleon} &$16\times$ & 0.11    &2.87   & 4.67   &12.08   &0.08  &0.18 &17.32 &0.36 &17.81 &0.56 \\
&LlamaGen~\cite{llamagen} &$16\times$ & 0.16     &4.28   & 5.41      &14.77  &0.07 &0.15 &11.17 &0.30 &18.22 &0.58  \\
&VAR~\cite{var} &$16\times$ & 1.24    &15.74  & 10.89    &34.19  &0.10 &0.23 &8.91 &0.24 &19.98 &0.63 \\
&MaskBit~\cite{weber2024maskbit} &$16\times$  &0.16 &2.54 &4.45   &10.85  &0.06  &0.11 &12.53 & 0.38 &18.07 &0.57 \\
&TokenFlow~\cite{tokenflow} &$16\times$   &0.28  &6.73   &6.41   &20.46  &0.07 &0.15  &9.09 &0.28 &18.74 &0.59 \\
&O-MAGVIT2~\cite{openmagvit2} &$16\times$  & 0.34    &7.52   & 6.46    &20.99  &0.08 &0.19 &8.51 &0.27 &19.05 &0.60 \\
&O-MAGVIT2(pretrain)~\cite{openmagvit2} &$16\times$  & 0.80    &10.58  & 9.59   &27.59   &0.08 &0.20  &8.39 &0.27 &19.33 &0.61  \\
&UniTok~\cite{ma2025unitok} &$16\times$  & \textbf{13.53}   &\textbf{44.59}  & \textbf{38.73}    &\textbf{65.84} &0.15 &0.35 &\textbf{7.82} &0.20 &21.15 &0.66 \\
&OmniTokenizer~\cite{OmniTokenizer}&$8\times$   &2.14  &20.63 &13.24  &39.14  &0.15  &0.37  & 9.26 & 0.30 &15.15 &0.59 \\
&LlamaGen(F8)~\cite{llamagen} &$8\times$  & 4.39   &29.41  & 19.69    &49.00  &0.17 &0.40 &8.65 &0.19 &21.50 &0.67  \\
&O-MAGVIT2(F8)~\cite{openmagvit2}  &$8\times$  & 9.33    &40.24  & 30.82    &59.97  &\textbf{0.23} &\textbf{0.48} &7.88 &\textbf{0.17} &\textbf{22.53} &\textbf{0.70}\\
\midrule
\multirow{5}{*}{Continuous} &DC-AE~\cite{xie2024sana}  &$32\times$ &1.42  &16.35 &10.95   &33.82   &0.10  &0.26 &12.88 & 0.23 &20.88 &0.65   \\
&VA-VAE~\cite{yao2025vavae} &$16\times$ &6.92    &37.04  &25.14     &56.32 &0.22 &0.49  &6.68 &0.16 &22.94 &0.70 \\
&SD-XL~\cite{sdxl}  &$8\times$ & 6.94   &34.21  & 25.03    &53.68  &0.18 &0.42 &7.60 &0.19 &22.52 &0.69  \\
&SD-3.5~\cite{sd3}  &$8\times$ & 36.26    &67.04  & 59.04    &80.58  &0.43 &0.70 &7.11 &0.13 &24.89 &0.75  \\
&FLUX.1-dev~\cite{flux2024}  &$8\times$ & \textbf{50.69}     &\textbf{75.91}  & \textbf{70.70}    &\textbf{86.42} &\textbf{0.52} &\textbf{0.76} &\textbf{6.42} &\textbf{0.11} &\textbf{25.50} &\textbf{0.77 }\\

\bottomrule
\end{tabular}}

\caption{\textbf{Performance of discrete and continuous tokenizer on TokBench.} `$_{s}$' and `$_{m}$'denote the average metrics for small-scale instances and all scales, respectively. In this table, we compute traditional metrics such as rFID across both the text set and face set. The `Factor' denotes the downsampling ratio in latent space, while `1D' indicates that images are encoded into one-dimension.
  } 
\label{tab:tokbench_256_results}
\vspace{-3mm}
\end{table*}

\subsection{Evaluation Setting}
In this section, we conduct comprehensive comparisons of existing classical continuous or discrete visual tokenizers on the proposed TokBench. 
% Since most tokenizers are trained at fixed resolutions but can generalize to higher resolutions while achieving better results, 
We evaluate image reconstruction quality at three resolutions: 256, 512, and 1024.
For each resolution, we first center-pad the original image into a square and then resize it to the target resolution. After reconstruction within the target resolution, we resize the image back to its original padding size and crop out the padded regions to obtain a reconstructed result matching the original resolution.
We additionally provide baseline results for each resolution by applying the same padding and resizing process without reconstruction, representing the theoretical upper limit at that resolution.
For video reconstruction, we conduct experiments under resolutions at 256 and 480. Notably, we resize the shorter edge of videos to these target lengths while padding both the longer edge and frame count to meet the required dimensions for tokenizers. After reconstruction, we crop out the padded regions and resize the videos back to their original resolutions. The reconstructed videos are then evaluated frame-by-frame using the same protocols as images.

Our evaluation framework demonstrates efficiency and lightweight characteristics. After the reconstruction of all images in TokBench, the complete calculation of T-ACC and F-Sim metrics for images requires \textbf{only 2GB of GPU memory and can be completed within 4 minutes} on a single RTX 4090 GPU. 
For evaluating all reconstructed videos, the process requires 2GB of GPU memory and approximately 30 minutes to complete, which can be reduced to 6 minutes through multi-GPU parallel processing.

\begin{table*}[!t]

\centering

\small
\setlength{\tabcolsep}{4pt}
\resizebox{\textwidth}{!}{\begin{tabular}{l l c  c c c c |c c c c   |c c c c }

\toprule

 \multirow{2}{*}{\textbf{Type}} & \multirow{2}{*}{\textbf{Method}} & \multirow{2}{*}{\textbf{Factor}} & \multicolumn{4}{c}{\bf T-ACC(\%){$\uparrow$}} & \multicolumn{4}{c}{\bf T-NED(\%){$\uparrow$}}  & \multirow{2}{*}{\bf rFID{$\downarrow$}} & \multirow{2}{*}{\bf LPIPS{$\downarrow$}} & \multirow{2}{*}{\bf PSNR{$\uparrow$}} & \multirow{2}{*}{\bf SSIM{$\uparrow$}} \\
 
\cmidrule(lr){4-7} \cmidrule(lr){8-11}
&& &Small &Medium &Large &Mean &Small &Medium &Large &Mean &&&& \\

\midrule
\multicolumn{15}{c}{\textit{Resolution: 256 $\times$ 256}} \\
\midrule
\cellcolor{lightblue} &\cellcolor{lightblue}Resize &\cellcolor{lightblue}$1\times$   &\cellcolor{lightblue}86.05    &\cellcolor{lightblue} 94.65    &\cellcolor{lightblue}98.37  &\cellcolor{lightblue}93.02  &\cellcolor{lightblue} 92.98    &\cellcolor{lightblue} 97.22    &\cellcolor{lightblue}99.38  &\cellcolor{lightblue}96.53 & \cellcolor{lightblue}5.66 & \cellcolor{lightblue}0.07 &\cellcolor{lightblue}25.40 &\cellcolor{lightblue}0.81 \\

\multirow{11}{*}{Discrete}  
&TiTok &$1D$ &0.05  &0.06  &0.17&0.09 & 3.04  & 4.07  &5.58&4.23 &18.41 & 0.50 &13.80 &0.50  \\
&FlexTok &$1D$ &  0.55   &  2.24   & 18.06&6.95 &  7.80   &  14.26  & 41.21&21.09  &11.01 & 0.31 &17.58 &0.61 \\
&VQGAN &$16\times$  &0.05     & 0.12     &3.14   &  1.10   & 4.34     & 5.33     &15.00  &8.22  &15.66 & 0.33 &17.17 &0.58  \\
&Chameleon &$16\times$ & 0.11     & 0.31     &8.19   &2.87   & 4.67     & 6.65     &24.91  &12.08 &17.60 & 0.33 &17.66 &0.59\\
&LlamaGen &$16\times$ & 0.16     & 0.44     &12.25  &4.28   & 5.41     & 7.50     &31.40  &14.77 &14.23 & 0.29 &18.04 &0.61  \\
&VAR  &$16\times$ & 1.24     & 6.72     &39.26  &15.74  & 10.89    & 26.26    &65.42  &34.19 &10.30 & 0.22 &19.74 &0.66  \\
&MaskBit  &$16\times$ &0.16   &  0.19   & 7.26 &2.54 &  4.45   &  5.72   & 22.37&10.85  &17.05 & 0.37 &17.90 &0.60 \\
&TokenFlow  &$16\times$  &0.28     &1.62     &18.29  &6.73   &6.41     &12.34    &42.64  &20.46 &11.04 & 0.26 &18.61 &0.62  \\
&O-MAGVIT2 &$16\times$ & 0.34     & 1.49     &20.73  &7.52   & 6.46     & 12.41    &44.10  &20.99  &10.18 & 0.25 &18.85 &0.63  \\
&O-MAGVIT2(pretrain) &$16\times$ & 0.80     & 3.17     &27.76  &10.58  & 9.59     & 19.15    &54.02  &27.59 & 9.83 & 0.24 &19.15 &0.64 \\
&UniTok &$16\times$ & \textbf{13.53}    & \textbf{42.87}    & \textbf{77.35}  & \textbf{44.59}  & \textbf{38.73}    & \textbf{68.51}    & \textbf{90.27}  & \textbf{65.84}  & 9.21 & 0.19 &20.58 &0.68  \\
&OmniTokenizer  &$8\times$  &  2.14   &  8.46   & 51.28&20.63&  13.24  &  30.50  & 73.67&39.14  &12.70 & 0.30 &14.73 &0.62  \\
&LlamaGen &$8\times$ & 4.39     & 17.86    &65.97  &29.41  & 19.69    & 44.56    &82.76  &49.00 &10.51 & 0.18 &20.85 &0.68 \\
&O-MAGVIT2 &$8\times$ & 9.33     & 34.16    &77.24  &40.24  & 30.82    & 59.89    &89.19  &59.97  & \textbf{8.99} & \textbf{0.16} & \textbf{21.71} & \textbf{0.71} \\
\midrule
\multirow{4}{*}{Continuous} &DC-AE &$32\times$  &  1.42   &  5.16   & 42.45&16.35&  10.95  &  24.06  & 66.45& 33.82 &14.61 & 0.22 &20.42 &0.67 
  \\
&VA-VAE &$16\times$  &6.92     &28.25    &75.96  &37.04  &25.14    &55.30    &88.52  &56.32 & 8.22 & 0.16 &21.94 &0.71 \\
&SD-XL &$8\times$ & 6.94     & 24.83    &70.85  &34.21  & 25.03    & 50.96    &85.03  &53.68 & 8.93 & 0.18 &21.69 &0.70  \\
&SD-3.5 &$8\times$ & 36.26    & 72.18    &92.68  &67.04  & 59.04    & 85.64    &97.06  &80.58 & 8.40 & 0.13 &23.46 &0.75 \\
&FLUX.1-dev &$8\times$ & \textbf{50.69}    & \textbf{82.14}    & \textbf{94.89}  & \textbf{75.91}  & \textbf{70.70}    & \textbf{90.67}    & \textbf{97.90}  & \textbf{86.42} & \textbf{7.19} & \textbf{0.12} & \textbf{23.93} & \textbf{0.76} \\

\midrule
\multicolumn{15}{c}{\textit{Resolution: 512 $\times$ 512}} \\
\midrule

\cellcolor{lightblue} &\cellcolor{lightblue}Resize &\cellcolor{lightblue}$1\times$  &\cellcolor{lightblue}  92.51    &\cellcolor{lightblue}  98.18    &\cellcolor{lightblue} 98.86  &\cellcolor{lightblue}96.52  &\cellcolor{lightblue}  96.25    &\cellcolor{lightblue}  99.24    &\cellcolor{lightblue} 99.64  &\cellcolor{lightblue}98.38 &\cellcolor{lightblue}0.26 & \cellcolor{lightblue}0.01 &\cellcolor{lightblue}29.80 &\cellcolor{lightblue}0.91 \\

\multirow{10}{*}{Discrete} &VQGAN &$16\times$ & 0.15    & 0.76    &17.45 &6.12  & 5.20    & 8.99    &37.77 &17.32  &6.87 & 0.19 &19.24 &0.65  \\
&Chameleon &$16\times$ &0.60     &2.67     &31.39  &11.55  &7.63     &17.82    &54.95  &26.80  &5.61 & 0.17 &19.81 &0.66 \\
&LlamaGen &$16\times$ &0.67     &3.93     &40.43  &15.01  &7.76     &20.17    &63.39  &30.44 &5.28 & 0.15 &20.21 &0.68  \\
&VAR &$16\times$ &3.71     &20.59    &63.62  &29.31  &18.01    &49.56    &82.44  &50.00  &3.78 & 0.12 &21.27 &0.73  \\
&TokenFlow &$16\times$ &1.06     &6.27     &44.88  &17.40  &10.00    &28.39    &68.07  &35.49  &5.14 & 0.15 &20.46 &0.68  \\
&O-MAGVIT2 &$16\times$ &1.40     &9.51     &54.04  &21.65  &10.79    &33.15    &74.94  &39.63 &3.65 & 0.13 &21.11 &0.71  \\
&O-MAGVIT2(pretrain) &$16\times$  &3.02     &16.25    &62.72  &27.33  &16.87    &44.50    &80.48  &47.28 &3.51 & 0.12 &21.54 &0.72 \\
&UniTok &$16\times$ &17.25    &51.86    &81.20  &50.10  &44.75    &76.28    &92.20  &71.08 &3.27 & 0.09 &22.54 &0.76 \\
&OmniTokenizer &$8\times$ &  6.21   &  38.33  & 82.91&42.48&  23.44  &  65.99  & 92.66&60.70 &5.67 & 0.20 &15.00 &0.67\\
&LlamaGen &$8\times$ &12.13    &56.66    &88.89  &52.56  &34.11    &77.45    &95.39  &68.99 &2.60 & \textbf{0.07} &23.46 &0.78  \\
&O-MAGVIT2 &$8\times$ & \textbf{20.66}    & \textbf{70.36}    & \textbf{90.66}  & \textbf{60.56}  & \textbf{46.60}    & \textbf{85.41}   & \textbf{96.00}  & \textbf{76.00} &\textbf{2.34} & \textbf{0.07} & \textbf{24.39} & \textbf{0.80} \\
\midrule
\multirow{4}{*}{Continuous} &DC-AE &$32\times$ &  5.31   &  30.10  & 79.33&38.25&  20.91  &  57.78  & 89.98&56.22 &2.33 & 0.09 &23.24 &0.76 \\
&VA-VAE &$16\times$   &12.72   &58.73    &88.43  &53.30  &34.80    &78.86    &95.30  &69.65 &2.23 & 0.07 &24.07 &0.79  \\
&SD-XL &$8\times$ &16.53    &62.87    &91.20  &56.86  &40.43    &80.83    &96.40  &72.55  &2.00 & 0.06 &24.67 &0.80   \\
&SD-3.5  &$8\times$ &56.55    &91.64    &97.33  &81.84  &75.56    &96.44    &98.91  &90.30  &1.33 & \textbf{0.03} &26.57 &0.85 \\
&FLUX.1-dev &$8\times$ &\textbf{70.29}    &\textbf{94.62}    &\textbf{98.02}  &\textbf{87.64}  &\textbf{84.67}    &\textbf{97.65}    & \textbf{99.26}  & \textbf{93.86} & \textbf{0.73} & \textbf{0.03} & \textbf{27.25} & \textbf{0.86} \\

\midrule
\multicolumn{15}{c}{\textit{Resolution: 1024 $\times$ 1024}} \\
\midrule

\cellcolor{lightblue} &\cellcolor{lightblue}Resize &\cellcolor{lightblue}$1\times$   &\cellcolor{lightblue}95.15     &\cellcolor{lightblue}98.39     &\cellcolor{lightblue}99.30   &\cellcolor{lightblue}97.61   &\cellcolor{lightblue}97.97    &\cellcolor{lightblue}99.33     &\cellcolor{lightblue}99.77   &\cellcolor{lightblue}99.02 &\cellcolor{lightblue}0.18 & \cellcolor{lightblue}0.01 & \cellcolor{lightblue}inf  &\cellcolor{lightblue}0.96  \\

\multirow{14}{*}{Discrete} 
&VQGAN  &$16\times$  & 0.76   &2.69    & 41.53 &15.00 & 7.90    &15.03   &63.47 &28.80  &4.03 & 0.11 &21.67 &0.74 \\
&Chameleon &$16\times$  & 3.00     &8.22     &59.33  &23.52  & 14.46    &29.14    &77.56  &40.39  &2.98 & 0.09 &22.33 &0.75 \\
&LlamaGen &$16\times$  & 3.30     &10.62    &67.63  &27.19  & 14.13    &33.02    &83.57  &43.58 &3.35 & 0.09 &22.74 &0.77 \\
&VAR &$16\times$  & 9.64     &30.08    &75.35  &38.36  & 29.07    &59.16    &89.51  &59.25 &4.85 & 0.10 &22.40 &0.79 \\
&TokenFlow &$16\times$  & 4.46     &14.86    &68.57  &29.30  & 18.62    &41.21    &84.43  &48.09 &3.34 & 0.09 &23.26 &0.78 \\
&O-MAGVIT2 &$16\times$  & 5.76     &20.74    &77.71  &34.74  & 19.42    &47.04    &89.63  &52.03 &2.47 & 0.07 &23.86 &0.80  \\
&O-MAGVIT2(pretrain) &$16\times$  & 9.08     &29.35    &79.77  &39.40  & 28.65    &57.43    &90.78  &58.95 &2.32 & 0.07 &24.46 &0.81 \\
&UniTok &$16\times$  & 26.90    &47.91    &74.29  &49.70  & 54.36    &72.48    &87.93  &71.59 &4.02 & 0.07 &24.22 &0.83 \\
&OmniTokenizer  &$8\times$  &   14.27  &  54.67  & 91.49&53.48&   36.63  &  76.92  & 96.53&70.02   &4.13 & 0.16 &15.30 &0.74  \\
&LlamaGen &$8\times$  & 25.42    &71.63    &94.61  &63.89  & 50.33    &86.45    &97.95  &78.24 & \textbf{1.74} & \textbf{0.04} &26.57 &0.86  \\
&O-MAGVIT2  &$8\times$   & \textbf{35.29}    & \textbf{78.91}    & \textbf{94.84}  & \textbf{69.68}  & \textbf{60.97}    & \textbf{90.36}    & \textbf{98.03}  & \textbf{83.12}  &2.21 & 0.06 & \textbf{27.07} & \textbf{0.88} \\
\midrule
\multirow{4}{*}{Continuous} &DC-AE  &$32\times$ & 15.32  &  48.36  & 92.72&52.14&   35.47  &  71.69  & 96.89&68.02 &1.11 & 0.04 &27.01 &0.85\\
&VA-VAE &$16\times$  & 25.14    &69.54    &93.84  &62.84  & 48.94    &85.17    &97.52  &77.21 &1.59 & 0.04 &27.31 &0.87 \\
&SD-XL &$8\times$  & 31.41    &75.83    &96.29  &67.84  & 56.60    &88.34    &98.52  &81.15 &1.01 & 0.03 &28.60 &0.88  \\
&SD-3.5 &$8\times$  & 74.88    &95.76    &98.50  &89.71  & 87.57    &98.15    &99.44  &95.05 &0.54 & 0.02 &29.80 &0.92  \\
&FLUX.1-dev &$8\times$   & \textbf{83.71}    & \textbf{96.83}    & \textbf{98.72}  & \textbf{93.09}  & \textbf{92.68}    & \textbf{98.69}    & \textbf{99.52}  & \textbf{96.96} & \textbf{0.41} & \textbf{0.01} & \textbf{30.55} & \textbf{0.94} \\

\bottomrule
·\end{tabular}}

\vspace{-2mm}
\caption{\textbf{Performance of discrete and continuous tokenizer on TokBench text-set.}   } 
\label{tab:text_detailed_results}
\vspace{-4mm}
\end{table*}

\subsection{Main Results}
We primarily evaluate performance at 256 resolution since most tokenizers are trained at this scale, with results presented in Table~\ref{tab:tokbench_256_results}. Most discrete tokenizers employ 16× downsampled spatial quantization (F16), while we additionally evaluate 8× downsampled (F8) variants of LlamaGen~\cite{llamagen} and Open-MAGVIT2~\cite{openmagvit2} tokenizers for comparison. At 256 resolution, discrete tokenizers demonstrate notably poor performance in reconstructing small-scale text and faces. UniTok's~\cite{ma2025unitok} multi-codebook design preserves finer details, achieving significantly superior text reconstruction compared to other tokenizers - even outperforming continuous-space VAEs from VA-VAE~\cite{yao2025vavae} and SDXL~\cite{sdxl}. For face reconstruction, UniTok also surpasses other F16 tokenizers. The higher-compression 1D tokenizer TiTok~\cite{titok} yields the weakest results for both text and face reconstruction. Notably, F8 tokenizers consistently outperform their F16 counterparts with identical architectures, while continuous VAEs from SD3.5~\cite{sd3} and FLUX~\cite{flux2024} achieve the highest scores.

Compared to conventional metrics (FID~\cite{FID}, LPIPS~\cite{LPIPS}, PSNR, SSIM~\cite{SSIM}), improved text reconstruction typically correlates with better scores. However, comparisons between UniTok vs. VA-VAE/SDXL and VAR~\cite{var} vs. Open-MAGVIT2 (pretrain) reveal contradictory trends. Moreover, FID and PSNR exhibit limited discriminative power for text/face reconstruction quality, even with substantial T-ACC and F-Sim variations, their metric gaps remain marginal in FID. This evidences existing metrics' inadequacy in comprehensively evaluating these specific reconstruction tasks.

\begin{figure}[tb]
  \centering
  \includegraphics[width=0.98 \linewidth]{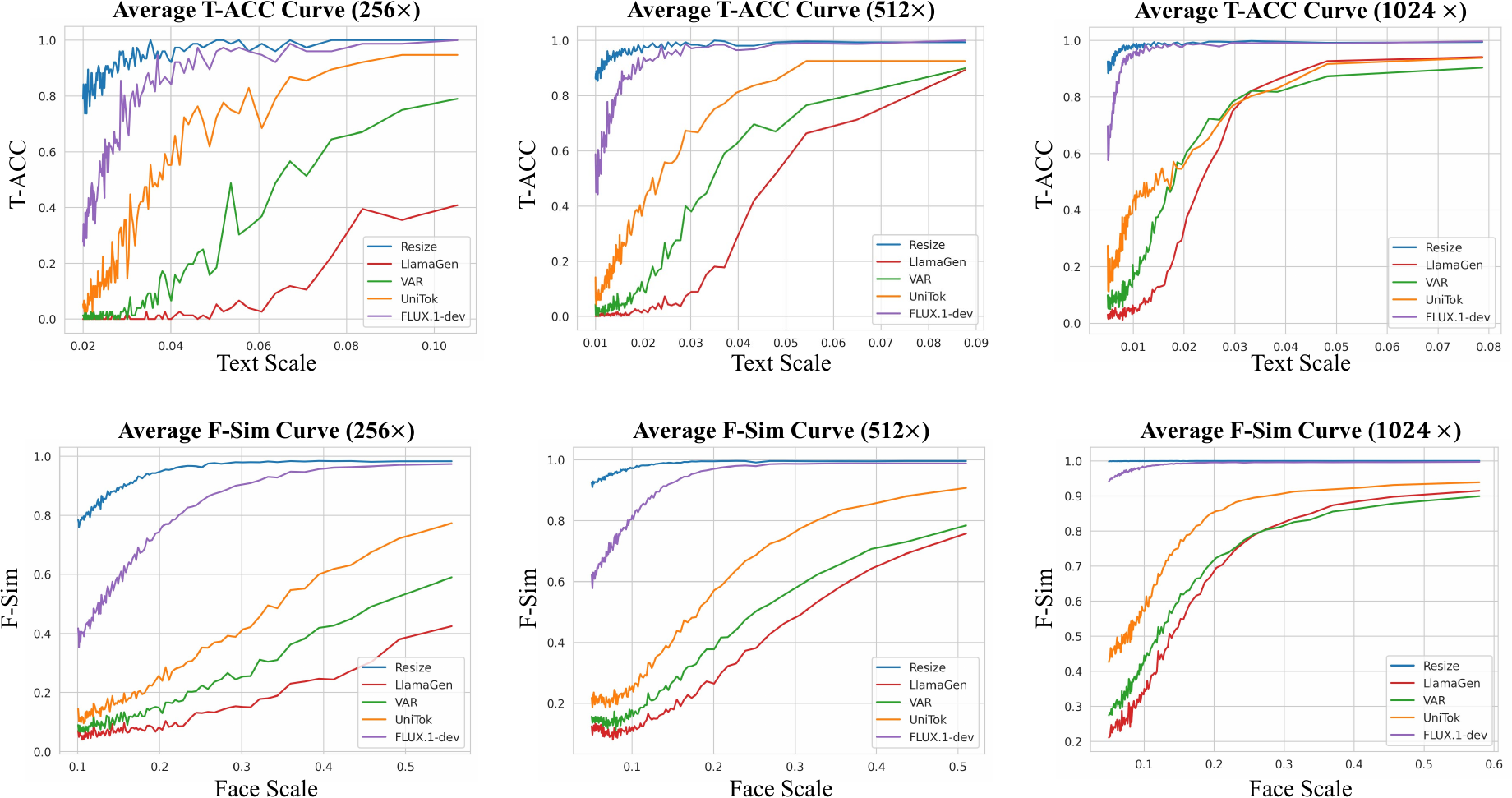}
   \vspace{-2mm}
  \caption{\textbf{T-ACC and F-Sim metrics across reconstruction resolutions versus target scales.} Smaller scales present greater challenges, and even the best-performing VAE show gap for improvement when compared to the ``resize'' upper bound.}
  \label{fig:curve}
\end{figure}

\subsection{Detail Evaluation for Text and Face} 
Table~\ref{tab:text_detailed_results} further presents the evaluation results of various tokenizers on text data across multiple resolutions. First, we observe that most tokenizers achieve progressively better performance with increasing resolution, even without being trained at 1024 resolution. Additionally, more discrepancies emerge between traditional metrics and T-ACC, as evidenced by cases like LlamaGen vs. TokenFlow at 512 resolution, UniTok vs. Open-MAGVIT2 at 1024 resolution, and LlamaGen(F8) vs. Open-MAGVIT2(F8) at 1024 resolution. These findings further validate the complementary value of our proposed metric to existing evaluation methods.

Notably, the performance gap between continuous and discrete tokenizers widens significantly with increasing resolution. At 1024 resolution, FLUX's VAE even achieves T-NED comparable to simple resizing. It's worth noting that since many original text images exceed 1024 pixels in size, even resizing cannot achieve 100\% T-ACC and T-NED. 
We further visualize the relationship between T-ACC/F-Sim metrics and instance scales across different resolutions in Figure~\ref{fig:curve}.
For small-scale objects, the performance gap between continuous and discrete tokenizers becomes more pronounced at higher resolutions.
Detailed evaluations on face data and the difficulty rating are provided in the supplementary materials.

\subsection{Video Tokenizers and VAEs}
\begin{table*}[!t]

\centering

\small
\setlength{\tabcolsep}{4pt}
\resizebox{\textwidth}{!}{\begin{tabular}{llc c c c c |c c c c   |c c c c }

\toprule

 \multirow{2}{*}{\textbf{Type}}    &\multirow{2}{*}{\textbf{Method}} &\multirow{2}{*}{\textbf{Factor}} & \multicolumn{4}{c}{\bf T-ACC(\%){$\uparrow$}} & \multicolumn{4}{c}{\bf T-NED(\%){$\uparrow$}}  & \multicolumn{4}{c}{\bf F-Sim{$\uparrow$}} \\
 
\cmidrule(lr){4-7} \cmidrule(lr){8-11} \cmidrule(lr){12-15}
&&&Small &Medium &Large &Mean &Small &Medium &Large &Mean  &Small&Medium &Large &Mean \\

\midrule
\multicolumn{15}{c}{\textit{Resolution: 256 $\times$ }} \\
\midrule
\cellcolor{lightblue} &\cellcolor{lightblue}Resize &\cellcolor{lightblue}$1\times1\times1$ &\cellcolor{lightblue} 76.09    &\cellcolor{lightblue} 92.14    &\cellcolor{lightblue} 96.18 &\cellcolor{lightblue} 88.14 &\cellcolor{lightblue} 85.77    &\cellcolor{lightblue} 95.79    &\cellcolor{lightblue} 98.32 &\cellcolor{lightblue} 93.29 & \cellcolor{lightblue}0.81 & \cellcolor{lightblue}0.91 &\cellcolor{lightblue}0.97 &\cellcolor{lightblue} 0.90 \\

\multirow{2}{*}{Discrete} 
&Cosmos-VAE~\cite{cosmos-tokenizer} &$4\times8\times8$  & 1.49 & 22.82 & 66.12& 30.14 & 7.76 & 44.61 & 77.15 & 43.18  & 0.29& 0.52     &0.76    & 0.52  \\
&Cosmos-VAE~\cite{cosmos-tokenizer} &$8\times16\times16$  &0.02 & 0.38 & 2.79 & 1.06 & 0.84 & 3.14 & 12.95 & 5.64  & 0.10& 0.13     &0.25    & 0.16   \\
\midrule
\multirow{5}{*}{Continuous}
&Cosmos-VAE~\cite{cosmos-tokenizer} &$4\times8\times8$ &5.80& 52.09 & 78.34 & 45.41 & 15.80 & 68.06 & 85.63 & 56.50  & 0.47& 0.72     &0.89    & 0.69   \\
&Hunyuan-Video~\cite{kong2024hunyuanvideo} &$4\times8\times8$ & \textbf{26.85}   &  69.12   & \textbf{87.47}  & 61.15  &  \textbf{45.55}  &  80.54  & \textbf{93.12} & \textbf{73.07} &  \textbf{0.60}&  \textbf{0.80}     & \textbf{0.92}    &  \textbf{0.77} \\
&CogVideoX~\cite{yang2024cogvideox} &$4\times8\times8$ &24.80  & \textbf{72.47}   & 86.34  & \textbf{61.21}  &  43.06  &  \textbf{82.29}  & 92.41 & 72.59 &  0.58&  0.78     & 0.91    &  0.76 \\
 &Cosmos-VAE~\cite{cosmos-tokenizer} &$8\times16\times16$  &0.45 & 6.23 & 48.99 & 18.56 & 3.25 & 24.62 & 64.08 & 30.65 & 0.21& 0.39     &0.65    & 0.42 \\
&Step-Video~\cite{ma2025step} &$8\times16\times16$ &  17.39    &  61.40    &  82.41 &  53.73 &  33.16  & 75.67   & 89.76  & 66.19 &  0.48&  0.69     & 0.86    &  0.67\\

\midrule
\multicolumn{15}{c}{\textit{Resolution: 480 $\times$ }} \\
\midrule

\cellcolor{lightblue} &\cellcolor{lightblue}Resize &\cellcolor{lightblue}$1\times1\times1$ 
 &\cellcolor{lightblue} 64.44    &\cellcolor{lightblue} 90.74    &\cellcolor{lightblue} 96.92  &\cellcolor{lightblue}84.04  &\cellcolor{lightblue} 77.71 &\cellcolor{lightblue} 95.72   &\cellcolor{lightblue} 98.57  &\cellcolor{lightblue} 90.67 & \cellcolor{lightblue}0.82 & \cellcolor{lightblue}0.89 &\cellcolor{lightblue}0.95 &\cellcolor{lightblue}0.89   \\

\multirow{2}{*}{Discrete}
&Cosmos-VAE~\cite{cosmos-tokenizer} &$4\times8\times8$  & 0.90 & 20.32 & 73.71 & 31.64 & 6.74 & 41.81 & 83.53 & 44.03 & 0.44 & 0.60 &0.80 & 0.61  \\
 &Cosmos-VAE~\cite{cosmos-tokenizer} &$8\times16\times16$ & 0.02 & 0.90 & 13.82 & 4.91 & 0.85 & 4.20 & 27.02 & 10.69 & 0.19 & 0.18 &0.31 & 0.23  \\
 
\midrule
\multirow{5}{*}{Continuous}
&Cosmos-VAE~\cite{cosmos-tokenizer} &$4\times8\times8$  & 5.30 & 46.80 & 86.82 & 46.31 & 14.99 & 64.63 & 92.20 & 57.27 & 0.60 & 0.77 &0.90 & 0.76  \\
&Hunyuan-Video~\cite{kong2024hunyuanvideo} &$4\times8\times8$ &\textbf{28.65}   &  64.49   & \textbf{91.83}  & 61.66  &  \textbf{44.43}  &  77.83  & \textbf{95.83} & \textbf{72.70} & \textbf{0.69}&\textbf{0.82}& \textbf{0.92}    &\textbf{0.81} \\
&CogVideoX~\cite{yang2024cogvideox} &$4\times8\times8$ &28.02  & \textbf{65.41}  &  91.71  & \textbf{61.71}  &  43.47  &  \textbf{78.24}  & 95.60 & 72.43 & 0.67 &0.80 & 0.91 &0.79   \\
&Cosmos-VAE~\cite{cosmos-tokenizer} &$8\times16\times16$ & 0.36 & 9.40 & 61.81 & 23.86 & 3.20 & 22.70 & 73.76 & 33.22 &  0.34 & 0.47 &0.71 & 0.51\\
&Step-Video~\cite{ma2025step} &$8\times16\times16$ &20.27    &  54.18   &  87.14 &  53.86 &  35.43  & 71.39   & 93.04  & 66.62   & 0.60&0.73& 0.86    &0.73\\

\bottomrule
\end{tabular}}
\caption{\textbf{Performance of video tokenizer on TokBench-Video.} The resolution refers specifically to the shorter edge of the videos, while maintaining the original aspect ratio throughout. The categorization into small, medium, and large scales is dynamically adjusted based on resolution.
 } 
\label{tab:video_results}
\vspace{-2mm}
\end{table*}
We evaluated video reconstruction quality at two standard resolutions (256 and 480) using a series of VAEs~\cite{cosmos-tokenizer} with identical architectures but varying compression ratios, along with three top-performing 3D causal VAEs from Step-Video~\cite{ma2025step}, Hunyuan-Video~\cite{kong2024hunyuanvideo}, and CogVideoX~\cite{yang2024cogvideox}, as shown in Table~\ref{tab:video_results}. 
Discrete video tokenizers remain understudied and demonstrate inferior performance. The Cosmos-VAE framework enables clear observation of the performance gap between discrete and continuous tokenizers under same architectural designs, while also revealing the impact of different compression factors.
While all $ 4\times 8 \times 8$ VAEs demonstrate effective video compression and reconstruction capabilities, their performance on small-scale text reconstruction still shows significant gaps compared to the theoretical upper bound (Resize). In contrast, face reconstruction achieves closer results to the theoretical upper bound, likely due to these VAEs' extensive facial data exposure during training.  
A comparison between the $8\times 16 \times 8$ Cosmos-VAE and Step-Video reveals that at identical compression ratios, Step-VAE demonstrates much more superior capabilities. Although its performance remains below that of Hunyuan-Video and CogVideoX's VAEs, it achieves an 8× compression ratio while maintaining highly efficient compression and reconstruction capabilities.

\begin{wraptable}{r}{8cm}
\centering
\vspace{-2ex}
\resizebox{1.0\linewidth}{!}{
\setlength{\tabcolsep}{2pt}
\begin{tabular}{llcccc}
\toprule 
\textbf{Method} & \textbf{Data}  &\textbf{T-ACC$_{s}\uparrow$} &\textbf{T-ACC$_{m}\uparrow$} &\textbf{T-NED$_{s}\uparrow$}  &\textbf{T-NED$_{m}\uparrow$}\\
    \midrule
    F16 & ImageNet &0.02 &2.96 &4.84 &12.11 \\
    F16 & ImageNet+Text  &0.09 &3.93 &5.19 &14.48 \\
    \midrule
    F8 & ImageNet    &2.99 &25.99 &16.25 &45.09  \\
    F8  & ImageNet+Text  &3.42 &27.51 &18.05 &47.36 \\
    \bottomrule
\end{tabular}}
\caption{Ablations on Training Data. While augmenting ImageNet with text-rich data yields performance improvements, the gains remain limited, indicating that model architecture design exerts a more substantial influence than training data composition.
}
\label{table:ablation}
% \vspace{-2mm}
\end{wraptable}

\subsection{Ablation of Training Data}
Since different tokenizers typically release weights trained on distinct datasets, we conduct ablation studies on training data to investigate its impact on text and face reconstruction performance. Following LlamaGen's~\cite{llamagen} training protocol, we augment the ImageNet~\cite{deng2009imagenet} dataset with an additional 230k text-rich images. We train both F16 and F8 VQGAN models for 400k steps on either the mixed dataset or the original ImageNet alone, then evaluate them on TokBench text set as shown in Table~\ref{table:ablation}. The results demonstrate that incorporating more text data indeed improves T-ACC and T-NED scores, though these improvements prove relatively marginal compared to architectural enhancements. This suggests that while training data influences text and face reconstruction quality, the tokenizer structural design remains the more critical factor. The detailed training data components are provided in the supplementary materials.

\vspace{-2mm}
\section{Limitation}
\vspace{-3mm}
In TokBench, the text reconstruction quality is judged based on the accuracy of text recognition.
Although the proposed metrics effectively reflect the reconstruction quality for these visual targets, they lack pixel-level probabilistic evaluation across the entire image. For instance, while text may be accurately reconstructed, distortions in contrast or saturation may occur, which our metrics cannot directly capture. Therefore, the proposed metrics should serve as a meaningful complement to commonly used metrics such as PSNR and FID, which evaluate reconstruction quality solely at the pixel level and statistics feature level respectively. 
\vspace{-2mm}
\section{Conclusion}
\vspace{-3mm}
In this work, we propose TokBench for evaluating the image and video compression quality of visual generative models, with targeted assessments of two challenging yet visually sensitive targets, text and human faces, which exhibit wide-scale distributions. Unlike conventional metrics focusing on pixel-level or global high-dimensional semantic information, we directly evaluate text readability and identity preservation, which are more perceptually critical to human observers. Leveraging mature toolchains, we achieve efficient and accurate assessment of reconstructed faces and text. Our experiments demonstrate that directly evaluating these elements serves as an effective complement to existing metrics, mitigating potential confusion or misleading results from previous approaches, thereby helping to ensure the upper bound of visual generation quality.

\clearpage
\bibliography{main}
\bibliographystyle{plain}

%%%%%%%%%%%%%%%%%%%%%%%%%%%%%%%%%%%%%%%%%%%%%%%%%%%%%%%%%%%%

\appendix

\section{Evaluation Setting}
% Technical appendices with additional results, figures, graphs and proofs may be submitted with the paper submission before the full submission deadline (see above), or as a separate PDF in the ZIP file below before the supplementary material deadline. There is no page limit for the technical appendices.

\subsection{Tokenizer Selection}
In this section, we detail the tokenizers used in our evaluation. For continuous-space compression VAEs, we employed VA-VAE~\cite{yao2025vavae} along with VAEs from SDXL~\cite{sdxl}, SD3.5~\cite{sd3}, and FLUX~\cite{flux2024} obtained from their HuggingFace models. For DC-AE, we employ the $32\times$ downsampling version with 32 latent dimensions as used in SANA~\cite{xie2024sana}.
For discrete VQVAEs and other discrete modeling approaches, we adopted the ImageNet-trained VQGAN~\cite{esser2021taming} model with a downsampling factor of f=16 and codebook dimensionality of 16,384 as our baseline. For LlamaGen~\cite{llamagen}, we utilized both F16 and F8 model variants. For VAR~\cite{var}, we selected the largest VAR-d36 model with 2.3B parameters.
The TiTok~\cite{titok} implementation used the TiTok-L-32 tokenizer, representing each image with 32 tokens. For Open-MAGVIT2~\cite{openmagvit2}, we evaluated both F16 and F8 models trained on ImageNet, along with an F16 model pretrained on 100M data featuring a codebook size of 262,144.
For MaskBit~\cite{weber2024maskbit}, we utilize the 12-bit variant. OmniTokenizer~\cite{OmniTokenizer} is implemented using the recommended imagenet\_k600 version, while FlexTok~\cite{flextok} adopts the version trained on the DFN dataset~\cite{dnfdataset}.

\begin{table*}[htb]

\centering

\small
\setlength{\tabcolsep}{4pt}
\resizebox{0.8\textwidth}{!}{\begin{tabular}{l l c c c c c  |c c c c }

\toprule

 \multirow{2}{*}{\textbf{Type}} & \multirow{2}{*}{\textbf{Method}} & \multirow{2}{*}{\textbf{Factor}} & \multicolumn{4}{c}{\bf Similarity{$\uparrow$}}   & \multirow{2}{*}{\bf rFID{$\downarrow$}} & \multirow{2}{*}{\bf LPIPS{$\downarrow$}} & \multirow{2}{*}{\bf PSNR{$\uparrow$}} & \multirow{2}{*}{\bf SSIM{$\uparrow$}} \\
 
\cmidrule(lr){4-7}  
&& &Small &Medium &Large &Mean  &&&& \\

\midrule
\multicolumn{11}{c}{\textit{Resolution: 256 $\times$ 256}} \\
\midrule
\cellcolor{lightblue} &\cellcolor{lightblue}Resize &\cellcolor{lightblue}$1\times$ &\cellcolor{lightblue}0.85     &\cellcolor{lightblue}0.97     &\cellcolor{lightblue}0.98    &\cellcolor{lightblue}0.93  &\cellcolor{lightblue} 7.83 &\cellcolor{lightblue} 0.05 &\cellcolor{lightblue}29.83 &\cellcolor{lightblue}0.87  \\

\multirow{14}{*}{Discrete} 
&TiTok  &$1D$  & 0.03     & 0.03     &0.05    & 0.04  &23.11 & 0.53 &13.31 &0.43\\
&FlexTok &$1D$  & 0.06     & 0.12     &0.25    & 0.15  &13.54 & 0.38 &17.18 &0.54 \\
&VQGAN  &$16\times$   & 0.05     & 0.08     &0.17    & 0.10  &18.08 & 0.38 &17.39 &0.52  \\
&Chameleon  &$16\times$  & 0.08     & 0.15     &0.30    & 0.18  &25.87 & 0.39 &17.94 &0.53 \\
&LlamaGen  &$16\times$     & 0.07     & 0.11     &0.26    & 0.15  &15.30 & 0.32 &18.38 &0.55  \\
&VAR  &$16\times$   & 0.10     & 0.20     &0.41    & 0.23     &13.11 & 0.25 &20.20 &0.61  \\
&MaskBit &$16\times$  & 0.06     & 0.09     &0.19    & 0.11  &15.92 & 0.39 &18.23 &0.55  \\
&TokenFlow  &$16\times$    & 0.07     & 0.13     &0.26    & 0.15  &13.43 & 0.30 &18.85 &0.56  \\
&O-MAGVIT2   &$16\times$    & 0.08  & 0.15   &0.34    & 0.19    &12.91 & 0.29 &19.24 &0.58   \\
&O-MAGVIT2(pretrain)  &$16\times$  & 0.08     & 0.16     &0.35    & 0.20  &12.92 & 0.29 &19.49 &0.59 \\
&UniTok  &$16\times$      & 0.15     & 0.32     &0.58    & 0.35   &11.25 & 0.21 &21.66 &0.65  \\
&OmniTokenizer  &$8\times$    & 0.15     & 0.34     &0.61    & 0.37 &12.06 & 0.31 &15.53 &0.56   \\
&LlamaGen  &$8\times$    & 0.17     & 0.38     &0.66    & 0.40  &12.01 & 0.20 &22.09 &0.66   \\
&O-MAGVIT2 &$8\times$  & 0.23     & 0.48     &0.74    & 0.48     &11.47 & 0.18 &23.27 &0.69  \\
\midrule
\multirow{5}{*}{Continuous} &DC-AE &$32\times$  & 0.10     & 0.21     &0.45    & 0.26  &17.58 & 0.25 &21.30 &0.62   \\
&VA-VAE  &$16\times$   & 0.22     & 0.48     &0.76    & 0.49  & 9.26 & 0.16 &23.85 &0.70 \\
&SD-XL  &$8\times$   & 0.18     & 0.40     &0.69    & 0.42   &11.19 & 0.20 &23.29 &0.68 \\
&SD-3.5  &$8\times$    & 0.43     & 0.76     &0.92    & 0.70   & 9.91 & 0.13 &26.20 &0.75 \\
&FLUX.1-dev  &$8\times$    & 0.52     & 0.83     &0.95    & 0.76   & 9.32 & 0.11 &26.94 &0.78 \\

\midrule
\multicolumn{11}{c}{\textit{Resolution: 512 $\times$ 512}} \\
\midrule

\cellcolor{lightblue} &\cellcolor{lightblue}Resize &\cellcolor{lightblue}$1\times$ &\cellcolor{lightblue}0.95    &\cellcolor{lightblue}0.99    &\cellcolor{lightblue}1.00     &\cellcolor{lightblue}0.98     &\cellcolor{lightblue}0.08 &\cellcolor{lightblue} 0.00 &\cellcolor{lightblue}37.34 &\cellcolor{lightblue}0.97 \\

\multirow{11}{*}{Discrete}  &VQGAN  &$16\times$   & 0.08      & 0.11     &0.37    & 0.19  &7.33 & 0.23 &20.42 &0.61 \\
&Chameleon  &$16\times$    & 0.13      & 0.21     &0.50    & 0.28   &6.62 & 0.22 &20.98 &0.61  \\
&LlamaGen  &$16\times$     & 0.11      & 0.17     &0.48    & 0.25   &5.28 & 0.18 &21.41 &0.65   \\
&VAR  &$16\times$   & 0.14      & 0.24     &0.57    & 0.32 &4.59 & 0.15 &22.16 &0.69  \\
&TokenFlow  &$16\times$   & 0.11      & 0.16     &0.45    & 0.24   &6.48 & 0.19 &21.39 &0.64  \\
&O-MAGVIT2  &$16\times$  & 0.13   & 0.22     &0.58    & 0.31  &4.67 & 0.16 &22.40 &0.67 \\
&O-MAGVIT2(pretrain)  &$16\times$      & 0.13      & 0.22     &0.57    & 0.31   &4.55 & 0.16 &22.66 &0.68     \\
&UniTok  &$16\times$      & 0.22      & 0.36     &0.74    & 0.44  &3.95 & 0.11 &24.34 &0.74 \\
&OmniTokenizer  &$8\times$    & 0.24      & 0.45     &0.80    & 0.50  &5.11 & 0.20 &15.93 &0.63 \\
&LlamaGen  &$8\times$   & 0.28      & 0.49     &0.83    & 0.53   &2.73 & 0.08 &25.49 &0.77  \\
&O-MAGVIT2 &$8\times$ & 0.35      & 0.58     &0.88    & 0.61  &2.80 & 0.07 &26.81 &0.80 \\
\midrule
\multirow{5}{*}{Continuous} &DC-AE  &$32\times$  & 0.16      & 0.29     &0.71    & 0.39    &2.81 & 0.11 &25.08 &0.73   \\ 
&VA-VAE  &$16\times$   & 0.31      & 0.54     &0.87    & 0.57   &2.41 & 0.07 &26.84 &0.79 \\
&SD-XL  &$8\times$   & 0.29      & 0.51     &0.87    & 0.55   &2.46 & 0.07 &27.14 &0.79   \\
&SD-3.5  &$8\times$   & 0.61      & 0.84     &0.98    & 0.81      &1.20 & 0.03 &30.06 &0.87  \\
&FLUX.1-dev  &$8\times$     & 0.71      & 0.89     &0.98    & 0.86    &0.71 & 0.02 &31.06 &0.90 \\

\midrule
\multicolumn{11}{c}{\textit{Resolution: 1024 $\times$ 1024}} \\
\midrule

\cellcolor{lightblue} &\cellcolor{lightblue}Resize &\cellcolor{lightblue}$1\times$  &\cellcolor{lightblue}1.0     &\cellcolor{lightblue}1.0     &\cellcolor{lightblue}1.0     &\cellcolor{lightblue}1.0    &\cellcolor{lightblue}0.01 &\cellcolor{lightblue} 0.00 &\cellcolor{lightblue} inf  &\cellcolor{lightblue}1.00   \\

\multirow{11}{*}{Discrete}  &VQGAN  &$16\times$   & 0.10      & 0.19      &0.47    & 0.25   &4.27 & 0.13 &23.98 &0.72\\
&Chameleon  &$16\times$     & 0.19      & 0.30      &0.58    & 0.36 &3.63 & 0.11 &24.54 &0.73  \\
&LlamaGen  &$16\times$    & 0.15      & 0.27      &0.57    & 0.33  &3.45 & 0.10 &24.77 &0.76 \\
&VAR  &$16\times$   & 0.23      & 0.34      &0.62    & 0.40   &6.21 & 0.12 &23.67 &0.77 \\
&TokenFlow  &$16\times$    & 0.14      & 0.26      &0.55    & 0.31 &4.63 & 0.10 &25.00 &0.76 \\
&O-MAGVIT2 &$16\times$   & 0.20      & 0.34      &0.66    & 0.40  &3.62 & 0.09 &26.01 &0.79  \\
&O-MAGVIT2(pretrain)  &$16\times$   & 0.21      & 0.34      &0.65    & 0.40 &3.48 & 0.09 &26.12 &0.79 \\
&UniTok  &$16\times$     & 0.33      & 0.50      &0.76    & 0.53    &3.78 & 0.07 &26.54 &0.84   \\
&OmniTokenizer  &$8\times$   & 0.40    & 0.60    & 0.82    & 0.61 &4.63 & 0.15 &16.00 &0.71\\
&LlamaGen  &$8\times$    & 0.49      & 0.66      &0.87    & 0.67 &2.11 & 0.04 &28.89 &0.87  \\
&O-MAGVIT2 &$8\times$  & 0.57      & 0.74      &0.91    & 0.74  &2.11 & 0.05 &29.92 &0.89  \\
\midrule
\multirow{5}{*}{Continuous} &DC-AE &$32\times$  &0.27    &0.45    &0.78    &0.50 &1.44 & 0.05 &29.48 &0.84 \\
&VA-VAE  &$16\times$   & 0.49      & 0.68      &0.89    & 0.69  &2.38 & 0.04 &30.69 &0.88   \\
&SD-XL  &$8\times$    & 0.50      & 0.69      &0.91    & 0.70 &1.25 & 0.03 &31.39 &0.89  \\
&SD-3.5  &$8\times$     & 0.86      & 0.94      &0.99    & 0.93   &0.42 & 0.01 &33.07 &0.96   \\
&FLUX.1-dev  &$8\times$   & 0.92      & 0.97      &0.99    & 0.96  &0.24 & 0.01 &33.61 &0.97  \\

\bottomrule
\end{tabular}}

\caption{\textbf{Performance of discrete and continuous tokenizer on TokBench face-set.}   } 
\label{tab:face_detailed_results}
\vspace{-2mm}
\end{table*}

\begin{wraptable}{r}{8cm}
\centering
\vspace{-2ex}
\resizebox{1.0\linewidth}{!}{
\setlength{\tabcolsep}{2pt}
\begin{tabular}{llcccc}
\toprule 
\textbf{Type} &\textbf{Cat.}  &\textbf{Res.} &\textbf{Small} &\textbf{Medium} &\textbf{Large}   \\
    \midrule
 \multirow{6}{*}{Image}   &\multirow{3}{*}{Text} &256  &$0.02\sim0.03$\quad\quad  &$0.03\sim0.04$\quad\quad  &$0.04\sim1.00$  \\
   &   &512    &$0.01\sim0.02$\quad\quad  &$0.02\sim0.03$\quad\quad  &$0.03\sim1.00$  \\
   &   &1024    &$0.005\sim0.01$\quad\quad  &$0.01\sim0.02$\quad\quad  &$0.02\sim1.00$  \\
    
    \cmidrule(lr){2-6}  
    &\multirow{3}{*}{Face} &256   &$0.10\sim0.20$\quad\quad  &$0.20\sim0.30$\quad\quad  &$0.30\sim1.00$\\
   &   &512   &$0.05\sim0.10$\quad\quad  &$0.10\sim0.20$\quad\quad  &$0.20\sim1.00$ \\
   &   &1024   &$0.02\sim0.05$\quad\quad  &$0.05\sim0.10$\quad\quad &$0.10\sim1.00$ \\
    \midrule
\multirow{4}{*}{Video}   &\multirow{2}{*}{Text}    &256    &$0.01\sim0.02$\quad\quad  &$0.02\sim0.03$\quad\quad  &$0.03\sim1.00$  \\
   &   &480    &$0.005\sim0.01$\quad\quad  &$0.01\sim0.02$\quad\quad  &$0.02\sim1.00$  \\
    \cmidrule(lr){2-6}  
    &\multirow{2}{*}{Face} &256     &$0.05\sim0.10$\quad\quad  &$0.10\sim0.20$\quad\quad  &$0.20\sim1.00$ \\
   &   &480   &$0.02\sim0.05$\quad\quad  &$0.05\sim0.10$\quad\quad &$0.10\sim1.00$ \\
    
    \bottomrule
\end{tabular}}
\caption{Difficulty Rating}
\label{table:difficulty}
\vspace{-1em}
\end{wraptable}
\subsection{Difficulty Rating}
As mentioned in Section 3, we classified different target instances into three difficulty levels based on their scales. For text reconstruction tasks, we theoretically assume that at least $5\times5$ pixels are required to represent a single character. Based on this lower bound, we filtered out targets that are theoretically unrepresentable at each reconstruction resolution. For instance, at 256 resolution, the minimum character scale equals $ 5 \div 256 \approx 0.02 $, so text instances with scales smaller than 0.02 are excluded from evaluation in this setting. As shown in the Table~\ref{table:difficulty}, we determined the scale lower bound for each resolution following this rule, and categorized all targets into small, medium, and large scales according to the distribution curve in Figure~\ref{fig:curve}.
For face evaluation, through visualization and performance analysis of `Resize' upper bound, we set 25 pixels as the minimum representation for recognizable faces. Based on this lower bound, we define minimum evaluable face scales for different resolutions, for instance, at 256 resolution, the minimum valid face scale is approximately  $ 25 \div 256 \approx 0.1 $.
For video evaluation, given that most videos follow a 16:9 aspect ratio and we resize the shorter edge to specified dimensions according to common evaluation standards, resulting in longer edges around 500 pixels, we adopted a more lenient rating strategy compared to image-level evaluation to accommodate these pre-processing differences.

\section{Detailed Comparison on Face Set}
Table~\ref{tab:face_detailed_results} presents a comprehensive evaluation of various tokenizers on the face set across multiple resolutions. First, most tokenizers achieve better performance as resolution increases. Since most face images do not exceed 1024 resolution, resizing to 1024 preserves nearly identical facial details, resulting in the highest possible similarity score of 1. At this resolution, both SD3.5 and FLUX VAEs achieve near-perfect performance (close to 1), while discrete VQVAEs only reach a maximum similarity of 0.5 for small-scale faces. This indicates a significant performance gap between discrete and continuous compression methods for small-scale objects, even at higher resolutions.
Furthermore, results degrade substantially at lower resolutions, demonstrating that facial features require higher resolutions to maintain quality.

\section{More Visualization}
\begin{figure}[tb]
  \centering
  \includegraphics[width=0.9 \linewidth]{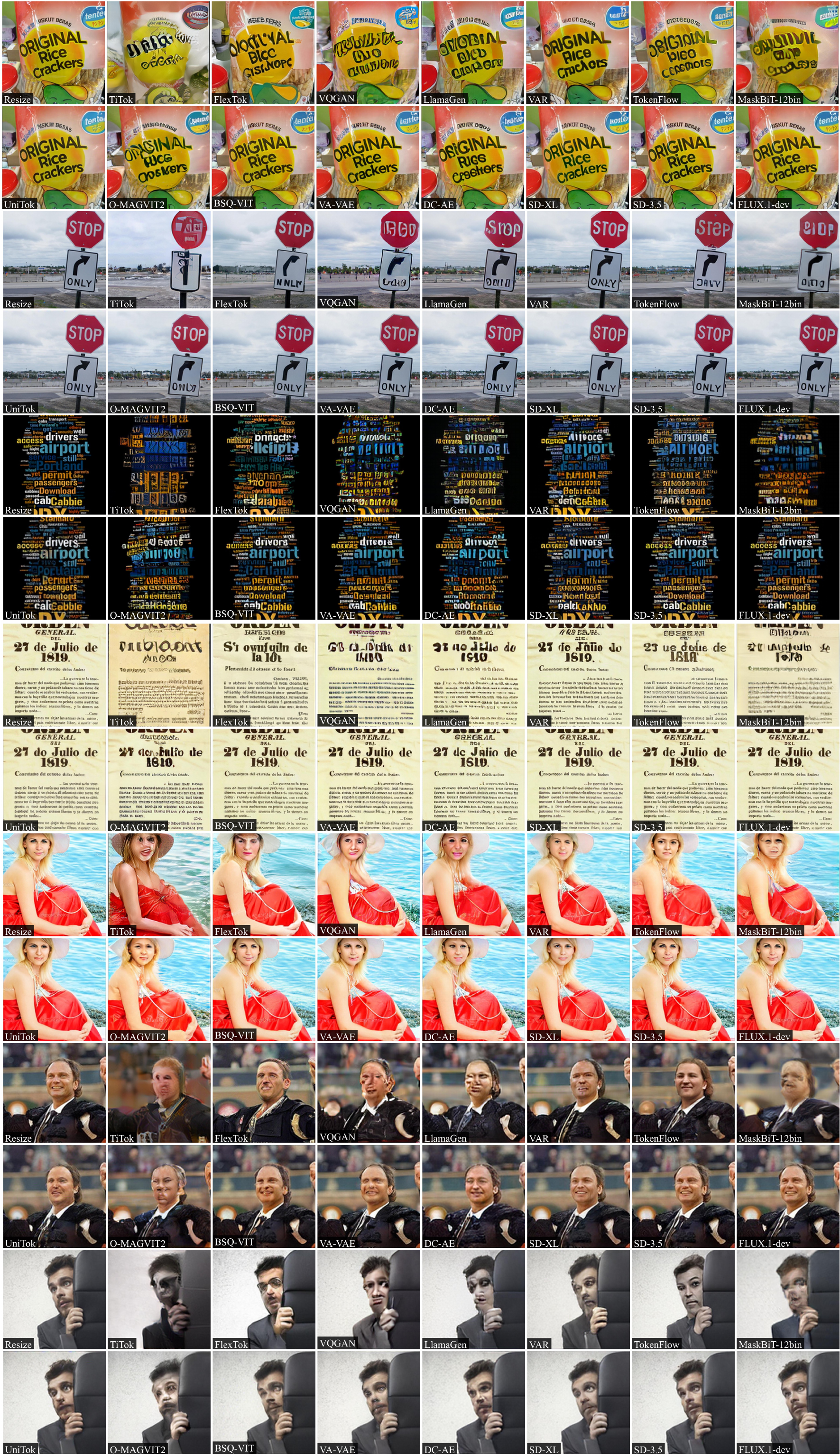}
   \vspace{-2mm}
  \caption{Visualization results of text and face reconstruction performance for different methods at 256 resolution. (Zoom in for better comparison.)}
  % \vspace{-4mm}
  \label{fig:visualize256}
\end{figure}

\begin{figure}[tb]
  \centering
    \includegraphics[width=0.9 \linewidth]{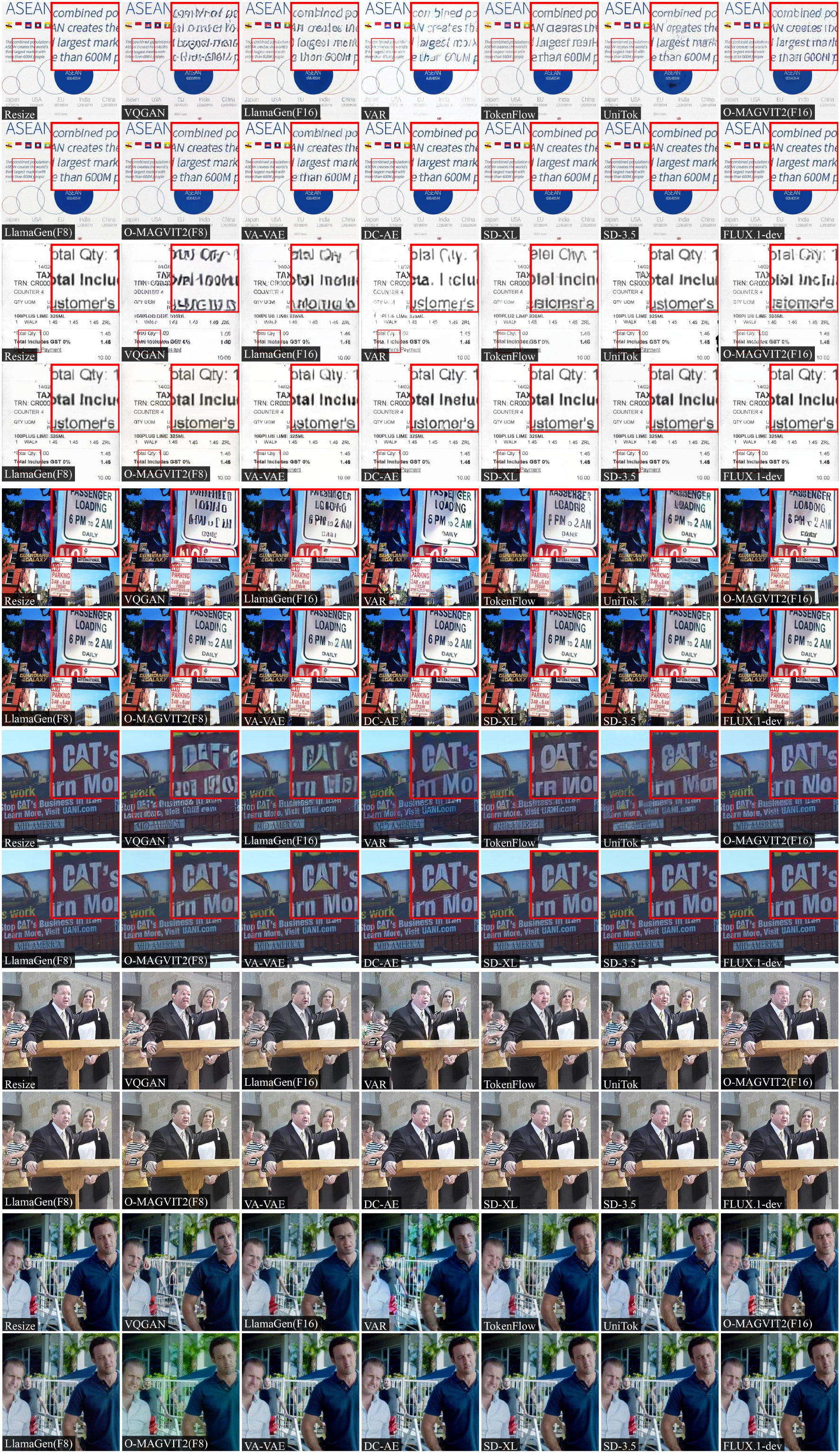}
   \vspace{-2mm}
  \caption{Visualization results of text and face reconstruction performance for different methods at 1024 resolution. (Zoom in for better comparison.)}
  % \vspace{-4mm}
  \label{fig:visualize1024}
\end{figure}

Tables ~\ref{fig:visualize256} and ~\ref{fig:visualize1024} present qualitative comparisons of reconstruction results from different methods at 256 and 1024 resolutions respectively. At 256 resolution, most discrete tokenizers fail to accurately reconstruct text and faces, while the high-compression DC-AE also performs poorly. In contrast, SD3.5 and FLUX VAEs demonstrate significantly better visual quality. At 1024 resolution, both VAEs and low-compression (F8) discrete tokenizers achieve satisfactory results, though F8 Open-MAGVIT2 exhibits noticeable color distortion, and F16 discrete tokenizers still struggle with small-scale objects.

\section{Ablation Setting}
In our ablation study examining the impact of text-rich training data augmentation. Following LlamaGen~\cite{llamagen}, we train VQGANs of F16 and F8 across two datasets. Our baseline implementation uses the ImageNet~\cite{deng2009imagenet} training set, for the ablation we supplement with 230,000 text-rich images sourced from the training sets of Synth150K~\cite{abcnet}, ICDAR 2017 MLT~\cite{mlt}, Total-Text~\cite{totaltext}, TextOCR~\cite{textocr}, CTW1500~\cite{ctw1500} and COCO-Text~\cite{veit2016coco}. The additional text images deviate from the evaluated data. Here, we only need the image rich in texts for training and no annotation is required. To ensure fair comparison, both training are executed for 400,000 iterations under identical conditions.

% % %%%%%%%%%%%%%%%%%%%%%%%%%%%%%%%%%%%%%%%%%%%%%%%%%%%%%%%%%%%%

\end{document}